\documentclass{article}

\usepackage{fancyvrb}

\usepackage[preprint]{neurips}

\usepackage[table]{xcolor}
\usepackage[most]{tcolorbox}
\definecolor{mycolor1}{HTML}{E2E7E6}
\definecolor{mycolor2}{HTML}{E4F1FF}
\usepackage[utf8]{inputenc} % allow utf-8 input
\usepackage[T1]{fontenc}    % use 8-bit T1 fonts
\usepackage{hyperref}       % hyperlinks
\usepackage{url}            % simple URL typesetting
\usepackage{booktabs}       % professional-quality tables
\usepackage{amsfonts}       % blackboard math symbols
\usepackage{nicefrac}       % compact symbols for 1/2, etc.
\usepackage{microtype}      % microtypography
\usepackage{xcolor}         % colors

\usepackage{amsmath}
\usepackage{graphicx}
\usepackage{pifont}
\usepackage{multirow}

\definecolor{mycolor}{rgb}{0.945, 0.055, 0.596}

\hypersetup{
  colorlinks=true,
  linkcolor=mycolor,
  urlcolor=mycolor
}

\title{AutoStudio: Crafting Consistent Subjects in Multi-turn Interactive Image Generation}

\author{%
  Junhao Cheng\textsuperscript{1}, Xi Lu\textsuperscript{1}, Hanhui Li\textsuperscript{1}, Khun Loun Zai\textsuperscript{1}, Baiqiao Yin\textsuperscript{1}, \\
\textbf{Yuhao Cheng\textsuperscript{2}, Yiqiang Yan\textsuperscript{2}, Xiaodan~Liang\textsuperscript{1}\thanks{Corresponding Author.}} \\ \\ \textsuperscript{1}Shenzhen Campus of Sun Yat-sen University, \textsuperscript{2}Lenovo Research \\ \\ \url{https://howe183.github.io/AutoStudio.io/}}

\begin{document}

\maketitle

\begin{figure*}[!h]
  \centering
  \includegraphics[width=1\textwidth]{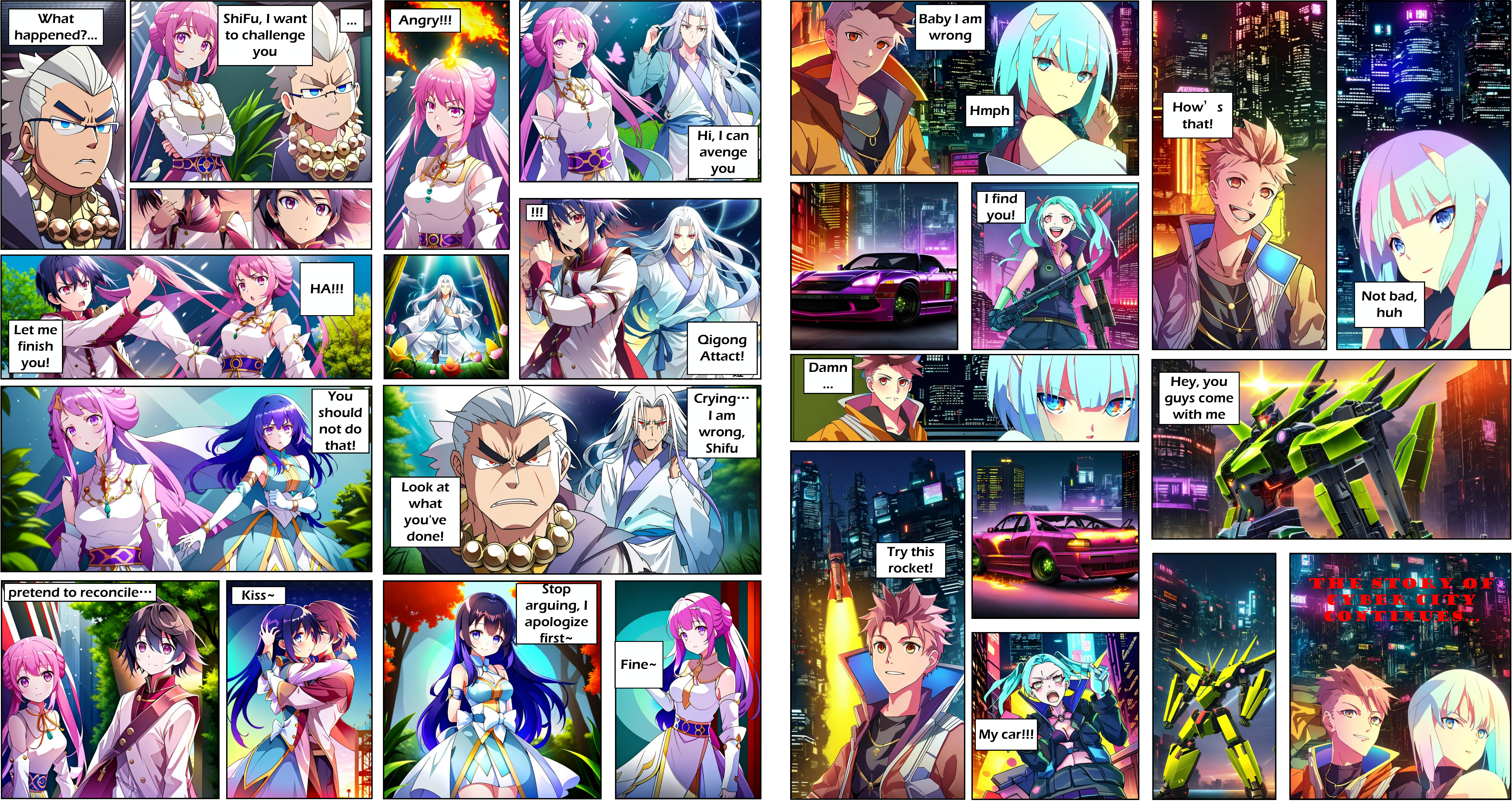}
  \caption{Two comic books generated by AutoStudio.}
  \label{fig: mamgaminx}
\end{figure*}

\begin{abstract}

As cutting-edge Text-to-Image (T2I) generation models already excel at producing remarkable single images, an even more challenging task, i.e., multi-turn interactive image generation begins to attract the attention of related research communities. This task requires models to interact with users over multiple turns to generate a coherent sequence of images. However, since users may switch subjects frequently, current efforts struggle to maintain subject consistency while generating diverse images. To address this issue, we introduce a training-free multi-agent framework called AutoStudio. AutoStudio employs three agents based on large language models (LLMs) to handle interactions, along with a stable diffusion (SD) based agent for generating high-quality images. Specifically, AutoStudio consists of (i) a \textbf{subject manager} to interpret interaction dialogues and manage the context of each subject, (ii) a \textbf{layout generator} to generate fine-grained bounding boxes to control subject locations, (iii) a \textbf{supervisor} to provide suggestions for layout refinements, and (iv) a \textbf{drawer} to complete image generation. Furthermore, we introduce a Parallel-UNet to replace the original UNet in the drawer, which employs two parallel cross-attention modules for exploiting subject-aware features. We also introduce a subject-initialized generation method to better preserve small subjects. Our AutoStudio hereby can generate a sequence of multi-subject images interactively and consistently. Extensive experiments on the public CMIGBench benchmark and human evaluations show that AutoStudio maintains multi-subject consistency across multiple turns well, and it also raises the state-of-the-art performance by 13.65\% in average Fréchet Inception Distance and 2.83\% in average character-character similarity. Our codes will be available at~\url{https://github.com/donahowe/AutoStudio.git}.

\end{abstract}

\section{Introduction}

\begin{figure*}[!b]
  \centering
  \includegraphics[width=1\textwidth]{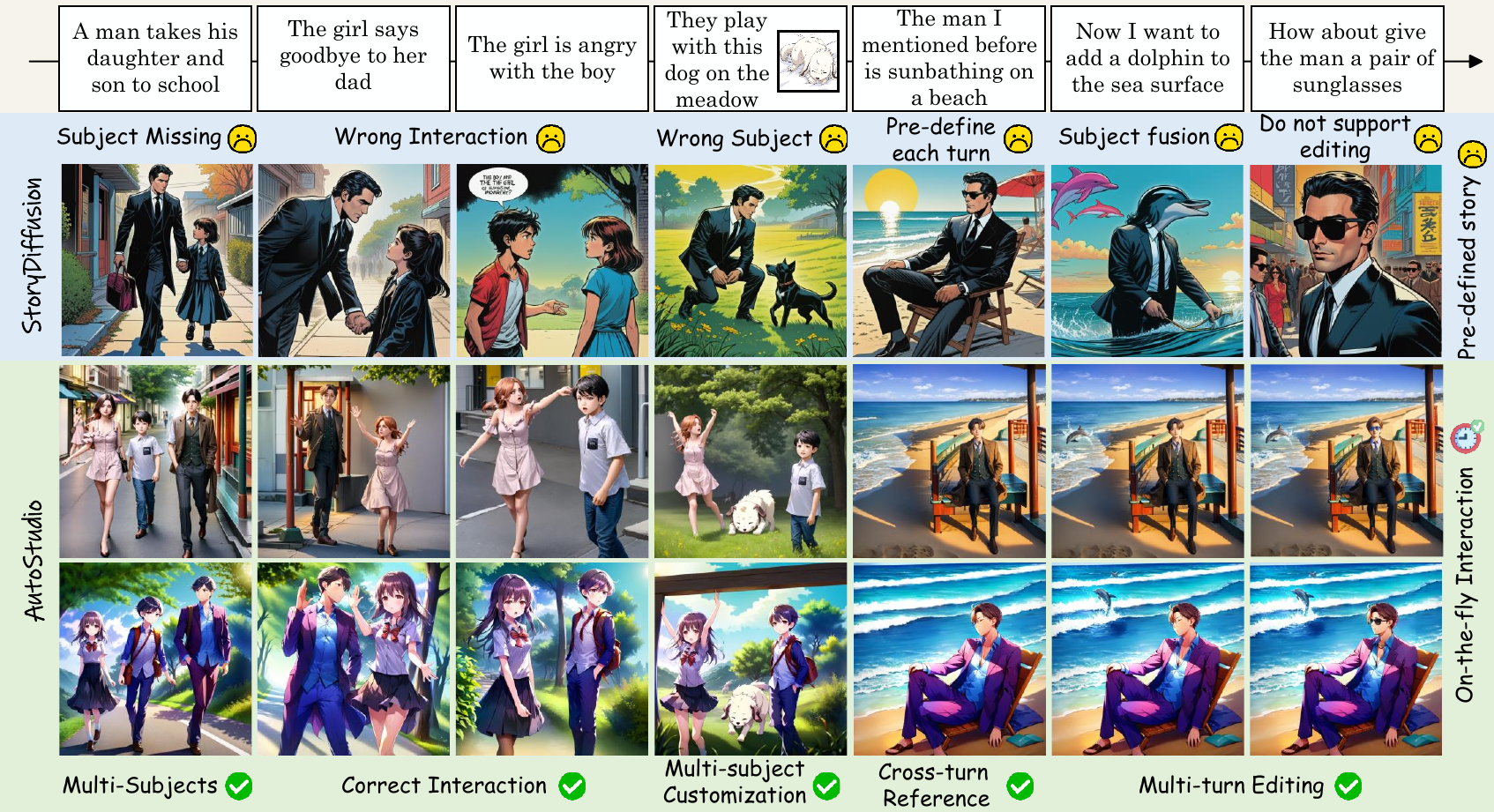}
  \caption{Visual examples of multi-turn interactive image generation tasks that can be achieved by AutoStudio while remaining challenging for other cutting-edge methods.}
  \label{fig: illustrate subject consistency}
\end{figure*}

%1 图像生成 -》 多轮图像生成
As cutting-edge T2I generation models have demonstrated exceptional capabilities in generating impressive individual images, there is a growing interest within the research communities regarding the more intricate undertaking of multi-turn interactive image generation~\cite{cheng2024theatergen,huang2024dialoggen, liu2023intelligent,Mini-DALLE3}. In real-world applications, users often require to generate a sequence of images in an interactive manner~\cite{cheng2024theatergen, Mini-DALLE3}, which encompass a wide range of tasks such as open-ended story generation and multi-turn editing with multiple subjects. However, current methods encounter difficulties in maintaining consistency across multiple subjects when faced with diverse user instructions, such as customization, editing, and extensive cross-turn references, as depicted in Figure~\ref{fig: mamgaminx} and~\ref{fig: illustrate subject consistency}. 

\begin{figure*}[!t]
  \centering
  \includegraphics[width=1\textwidth]{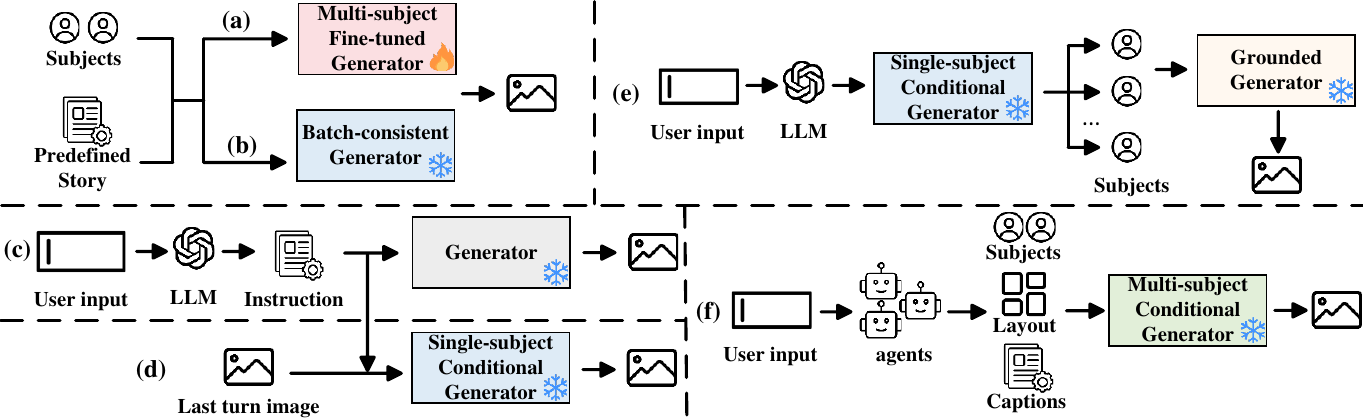}
  \caption{Architecture comparison  between AutoStudio (f) and other models, including (a) AutoStory, (b) StoryDiffusion, (c) Mini-Gemini, (d) Mini DaLLE·3, and (e) TheaterGen.}
  \label{fig: architecture compare}
\end{figure*}

As described in Figure~\ref{fig: architecture compare} and Table~\ref{tab: compare methods}, the architectures of previous models all have certain drawbacks. AutoStory~\cite{AutoStory} and TaleCrafter~\cite{TaleCrafter} fine-tune diffusion models with Low-Rank Adaptation (LoRA)~\cite{hu2021lora} to pre-define the characteristics of each subject, which diminishes the diversity of subjects. StoryDiffusion~\cite{zhou2024storydiffusion} requires a complete story to generate multiple images simultaneously, which sacrifices the flexibility of on-the-fly interaction and individual image editing. Moreover, generating all images at once also yields inferior results, as shown in Figure~\ref{fig: illustrate subject consistency}. Mini-Gemini~\cite{team2023gemini} utilize large multi-modal models as a router to comprehend and expand prompts to maintain contextual consistency. Additionally, Mini DALLE·3~\cite{Mini-DALLE3} takes into consideration the most recent image as a reference. However, the limited ability of the T2I model to understand complex prompts resulted in poor consistency. TheaterGen~\cite{cheng2024theatergen} generates each subject individually and merges them with ControlNet~\cite{controlnet}, which omits interactions among subjects and may yield unnatural results.

To tackle these issues, we introduce \textbf{AutoStudio}, a multi-agent training-free framework featuring four specially customized agents that employ off-the-shelf models to engage in on-the-fly interaction with users. Our intention is to introduce a versatile and scalable framework with multi-agent collaboration, allowing us to incorporate any desired LLM architecture and diffusion backbones into the framework to meet the diverse multi-turn generative requirements of users.

Specifically, AutoStudio consists of three LLM-based agents: (i) a \textbf{subject manager} interprets the dialogue, identifies different subjects, and assigns them with proper context; (ii) a \textbf{layout generator} generates part-level bounding boxes for each subject to control subject locations; (iii) a \textbf{supervisor} provides suggestions to the layout generator for layout refinement and correction. Finally, (iv) a \textbf{drawer}, which is based on Stable Diffusion (SD) \cite{sd}, completes image generation conditioned on the refined layout. 

Moreover, we introduce a Parallel-UNet (P-UNet) in the drawer, which has a novel architecture that utilizes two parallel cross-attention modules to enhance latent subject features with text and image embeddings separately. To further address the limitations of SD in understanding long prompts, as well as the issue of missing and mistakenly fused subjects during the generation process, we introduce a subject-initialized generation method in the drawer. 

With the above four agents collaborating closely, AutoStudio demonstrates significant advantages in multi-turn interactive image generation with the collaboration of multiple agents. Quantitative results on CMIGBench~\cite{cheng2024theatergen} show that AutoStudio raises the performance bar of the previous state-of-the-art TheaterGen method by 13.65\% in average Fréchet Inception Distance and 2.83\% in average character-character similarity. We also demonstrate the superiority of AutoStudio through human evaluation and qualitative analysis.

In summary, our contributions are as follows:
\begin{enumerate}
    \item We propose a training-free multi-agent framework called AutoStudio. This framework stands out for its ability to maintain multi-subject consistency in on-the-fly multi-turn interactions with users, enabling it to accomplish various tasks such as open-ended story/manga book generation and multi-turn editing.

    \item We propose a novel parallel UNet architecture with dual cross-attention modules to better exploit and fuse subject-aware text features and image features.

    \item We introduce a subject-initialized generation process to achieve finer controls of subject locations, which also alleviates the issues of missing subjects and erroneous subject fusions.

    \item AutoStudio outperforms the existing methods by large margins on the CMIGBench benchmark for multi-turn interactive image generation.

\end{enumerate}

\begin{table}[t!]
    \caption{Comparison between existing multi-turn image generation methods and AutoStudio.}
    \label{tab: compare methods}
    \centering
\resizebox{1\textwidth}{!}{
\begin{tabular}{lcccccc}
\toprule
 Methods/ task & On-the-fly & Multi-subject Interaction & Multi-subject Consistency & Multi-turn Editing \\
\midrule
AutoStory~\cite{AutoStory}, TaleCrafter~\cite{TaleCrafter} & \ding{55} & Limited & \checkmark & \ding{55} \\
SEED-LLAMA~\cite{SEED}, Mini-Gemini~\cite{team2023gemini} & \checkmark  & \checkmark  & Limited & \ding{55} \\
Mini DALL·E 3~\cite{Mini-DALLE3}, Intelligent Grimm~\cite{liu2023intelligent} & \checkmark & Limited & Limited & \ding{55} \\
Theatergen~\cite{cheng2024theatergen} & \checkmark & Limited & \checkmark & Limited \\
StoryDiffusion~\cite{zhou2024storydiffusion} & \ding{55} & Limited & Limited & \ding{55} \\

\textbf{AutoStudio (ours)} & \checkmark  & \checkmark  & \checkmark  & \checkmark \\
\bottomrule
\end{tabular}
}
\end{table}

\section{Related Works}

\subsection{Text-to-image Generation}
Text-to-image generation is a widely studied field, with notable methods such as Variational AutoEncoder (VAE) \cite{tibebu2022text}, flow-based models \cite{lu2020structured}, and Generative Adversarial Networks (GANs) \cite{GAN, GANsu, styleGAN, styleGAN2}. Recently, diffusion models \cite{ddpm, sd, sdxl, ipadapter, controlnet, DALL·E2, DALL·E3, avrahami2023chosen,cheng2025animegamer,cheng2024theatergen,luo2025object} have gained significant attention within the research community, especially the Stable Diffusion family \cite{sd, sdxl}. We also employ SD to implement the drawer in AutoStudio and augment it with to the ability to maintain features of multiple subjects.

\subsection{Multi-turn Interactive Image Generation}

Mini DALL·E 3~\cite{Mini-DALLE3} first introduced the concept of multi-turn interactive image generation. Mini-Gemini~\cite{team2023gemini} maintain consistency among subjects by translating and transforming prompts using LLM-based Rooters. Theatergen utilizes LLMs for character management and individual customization. Intelligent Grimm~\cite{liu2023intelligent} incorporates a visual language context module to extract information from previous-turn images. StoryDiffusion~\cite{zhou2024storydiffusion} introduces a hot-pluggable attention module to incorporate role features. In contrast to existing methods, AutoStudio leverages the close collaboration of multiple agents to maintain subject consistency and generate high-quality images.

\subsection{Multi-Agents Collaboration}
Large models have revolutionized multi-agent systems, demonstrating their versatility and effectiveness across various applications~\cite{Qian_Cong_Yang_Chen_Su_Xu_Liu_Sun_2023, tang2024prioritizing,Hong_Zheng_,Park_O}. While collaborative agents outperform individual agents in tackling dynamic and complex tasks~\cite{Guo_Chen_Wang,wu2023autogen,wu2023autogen,Mora}. Inspired by existing works, we extend the concept of collaborative agents to enable multi-turn interactive image generation tasks.

\section{Methodology}

In this section, we introduce the details of AutoStudio for multi-turn interactive image generation. We begin by providing our task formulation and the overall multi-agent architecture of AutoStudio in Section~\ref{sec: Architecture}. We then introduce the core agents of AutoStudio for extracting well-organized drawing prompts from multi-turn user conversations (Sec. \ref{sec: llm_agent}) and generating high-quality images with multi-subject consistency (Sec. \ref{sec: drawer}).

\subsection{Overall Framework}
\label{sec: Architecture}

\textbf{Problem Formulation}. We focus on the challenging multi-turn interactive image generation task in this paper. Let $K \gg 1 $ denote the maximum possible number of interaction turns and $k = 2, ..., K$ be an arbitrary turn. Given the prompt of the $k$-th turn $p_k$, a set of history prompts $\mathcal{P} = \{p_1, ... , p_{k-1}\}$ and their corresponding synthesized images ${\mathcal{I}}=\{I_1, ..., I_{k-1}\}$, our goal is to generate the image of the current-turn $I_k$, in which subjects are consistent to those in ${\mathcal{I}}$. Assume there are $n$ unique subjects in ${\mathcal{I}}$. To facilitate fine-grained subject modifications and cross-subject interactions, we assume that each subject is composed of up to $m$ components. We construct a subject database $\mathcal{D}$ to distinguish and keep track of these subjects as follows:
\begin{equation}
\mathcal{D}=\{[\mathcal{S}_i,\mathcal{ID}_i, \{(\mathcal{S}_{i,j}, \mathcal{ID}_{i,j})|j=1,...,m\}]|i=1,...,n\},
\end{equation}
where $\mathcal{ID}_i$ and $\mathcal{ID}_{i,j}$ denote the unique identifier of the $i$-th subject and its $j$-th component. Unlike \cite{cheng2024theatergen} that stores subject images, $\mathcal{S}_i$ and $\mathcal{S}_{i,j}$ are image features of subject $i$ and its corresponding component, and hence we avoid unnecessary repetitive image encoding processes. The proposed AutoStudio can be formulated as follows:
\begin{equation}
 I_k = \Phi_{\text{AutoStudio}}(p_k, \mathcal{P}, \mathcal{D}).
 \label{eq:formulation}
\end{equation}
Note that $I_k$ in Eq. (\ref{eq:formulation}) is determined only by the prompt of the current turn and results of previous turns, which differs from previous methods \cite{AutoStory,TaleCrafter,zhou2024storydiffusion} that all prompts are provided in advance.

\begin{figure*}[!t]
  \centering
  \includegraphics[width=1\textwidth]{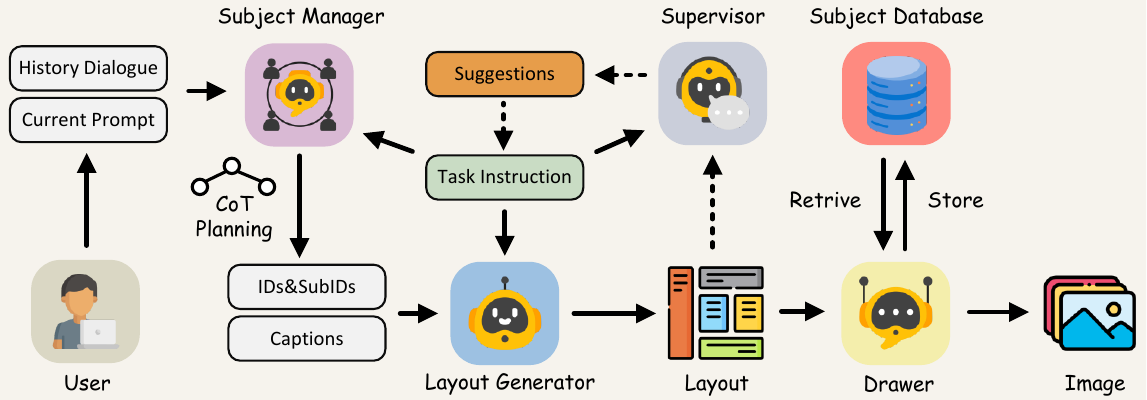}
  \caption{Overall structure of AutoStudio. AutoStudio leverages four agents and a subject database to complete multi-turn multi-subject interactive image generation: (i) A subject manager interprets user dialogues; (ii) A layout generator provides layout; (iii) A supervisor provides suggestions for layout refinement; (iv) A drawer generates images given refined layouts and the subject database.} 
  \label{fig: structure}
\end{figure*}

\textbf{Multi-agent Framework}. The overall framework of AutoStudio is shown in Figure \ref{fig: structure}, which consists of three LLM-based agents and a T2I drawer. Here, we use the three LLM-based agents for converting user prompts into drawing instructions, because recent research \cite{Guo_Chen_Wang, wei2022chain, Mora} suggests that step-by-step guidance boosts the performance of LLMs significantly. Specifically, we first introduce a subject manager $\mathcal{A}_{Manager}$ that not only assigns IDs to subjects and their components but also converts user prompts into drawing captions. These captions are then processed by the layout generator $\mathcal{A}_{Layout}$ to yield a coarse layout, which contains the bounding boxes and information of each subject and its components. To remedy irrational intra-and inter-subject spatial relationships and refine the coarse layout, a supervisor $\mathcal{A}_{Supervisor}$ is introduced. This supervisor takes the coarse layout as input and provides suggestions to the layout generator. In this way, $\mathcal{A}_{Supervisor}$ and $\mathcal{A}_{Layout}$ collaborate closely and form a closed-loop process for layout refinement. Moreover, we also define a set of task introductions to guide these three LLM-based agents to generate responses with proper formats. Finally, given the refined layout and subject information retrieved from $\mathcal{D}$, the drawer $\mathcal{A}_{drawer}$ can generate an image that is aligned with the layout well and contains consistent subjects. 

\subsection{Multi-turn Interaction Interpretation}
\label{sec: llm_agent}

\textbf{Subject Manager}.
Due to complex referential relationships and generation requirements in user prompts \cite{huang2024dialoggen,Mini-DALLE3,cheng2024theatergen}, feeding them directly into the drawer is inappropriate. Hence, we adopt a divide-and-conquer strategy that first utilizes $\mathcal{A}_{Manager}$ to process prompts and identity each subject. Let $\mathcal{O}_{Manager}^k$ denote the output of $\mathcal{A}_{Manager}$ w.r.t. the prompt of the $k$-th turn $p_k$, we consider to generate $\mathcal{O}_{Manager}^k$ by feeding $p_k$ along with all its previous prompts and corresponding outputs from $\mathcal{A}_{Manager}$ as follows:
\begin{equation}
 \mathcal{O}_{Manager}^k = \mathcal{A}_{Manager}( p_k, \{(\mathcal{O}_{Manager}^i, p_i)|i=1,...,k-1\}).
\end{equation}

To ensure $\mathcal{O}_{Manager}^k$ assigns the proper identifier and caption for each subject (and its components), we utilize chain-of-though prompting \cite{wei2022chain} with a pre-defined task instruction: ``\emph{Generate ID first, then assign sub-IDs to its important features.}" This allows us to obtain $\mathcal{O}_{Manager}^k$ with the following format:
\begin{equation}
 \mathcal{O}_{Manager}^k := \{c_{glb},c_{bg},\{[c_i,\mathcal{ID}_i, \{(c_{i,j}, \mathcal{ID}_{i,j})|j=1,...,m\}]|i=1,...,n\} \},
 \label{eq: o_manager}
\end{equation}
where $c_{glb},c_{bg}, c_i, c_{i,j}$ denote the global caption, background caption, the caption for subject $i$ and its component $j$, respectively. We assign every subject a unique $\mathcal{ID}$ that remains unchanged in the whole dialogue so that we can retrieve different subjects across multiple turns effectively.

\textbf{Layout Generator}.
The role of $\mathcal{A}_{Layout}$ is to generate a bounding box $b$ for each subject/component defined by $\mathcal{O}_{Manager}^k$, which can be expressed as follows:
\begin{equation}
 \mathcal{O}_{Layout}^k = \mathcal{A}_{Layout}(S, \mathcal{O}_{Manager}^k),
\end{equation}
where $S$ is the expected size of the generated image. Each generated bounding box $b$ is represented by the coordinates of its top-left corner, width, and height, namely, $b = [x_{\text{left}}, y_{\text{top}}, \text{width}, \text{height}]$. For the convenience of subsequent image generation and layout refinement, we also maintain subject information in $\mathcal{O}_{Layout}^k$. Hence, the format of $\mathcal{O}_{Layout}^k$ is defined as,
\begin{equation}
 \mathcal{O}_{Layout}^k := [\mathcal{O}_{Layout}^k, \{(b_i, \{b_{i,j}|j=1,...,m\})| i=1,..,,n\}]. 
\end{equation}

\textbf{Supervisor}.
Although LLMs possess strong interpretation capabilities, generating correct and reasonable bounding boxes for multiple objects at once is still challenging \cite{zheng2023layoutdiffusion,cheng2024theatergen,lin2024layoutprompter, anonymous2024visdiahalbench}. Hence we introduce $\mathcal{A}_{Supervisor}$ to provide suggestions for improving layouts. Similar to the above two agents, this process can be defined as, 
\begin{equation}
\mathcal{O}_{Supervisor}^k = \mathcal{A}_{Supervisor}(\mathcal{O}_{Layout}^k).
\end{equation}
Here, $\mathcal{O}_{Supervisor}^k$ contains multiple suggestions (e.g. "The hat should be positioned on top of the person's head."). The generated suggestions will be provided as feedback to  $\mathcal{A}_{Layout}$ for generating the final layout, which can be expressed as:
\begin{equation}
 \hat{\mathcal{O}}_{Layout}^k = \mathcal{A}_{Layout}(\mathcal{O}_{Supervisor}^k, \mathcal{O}_{Layout}^k).
\end{equation}
Details of the task instructions for $\mathcal{A}_{Manager}$, $\mathcal{A}_{Layout}$, and $\mathcal{A}_{Supervisor}$ can be found in the Appendix. We hereby obtain comprehensive captions and refined layouts regarding the target image of the current turn. These pieces of information are fed into the drawer $\mathcal{A}_{Drawer}$ to generate images with multi-subject consistency.

\subsection{Image Generation with Multi-subject Consistency}
\label{sec: drawer}

\begin{figure*}[!t]
  \centering
  \includegraphics[width=1\textwidth]{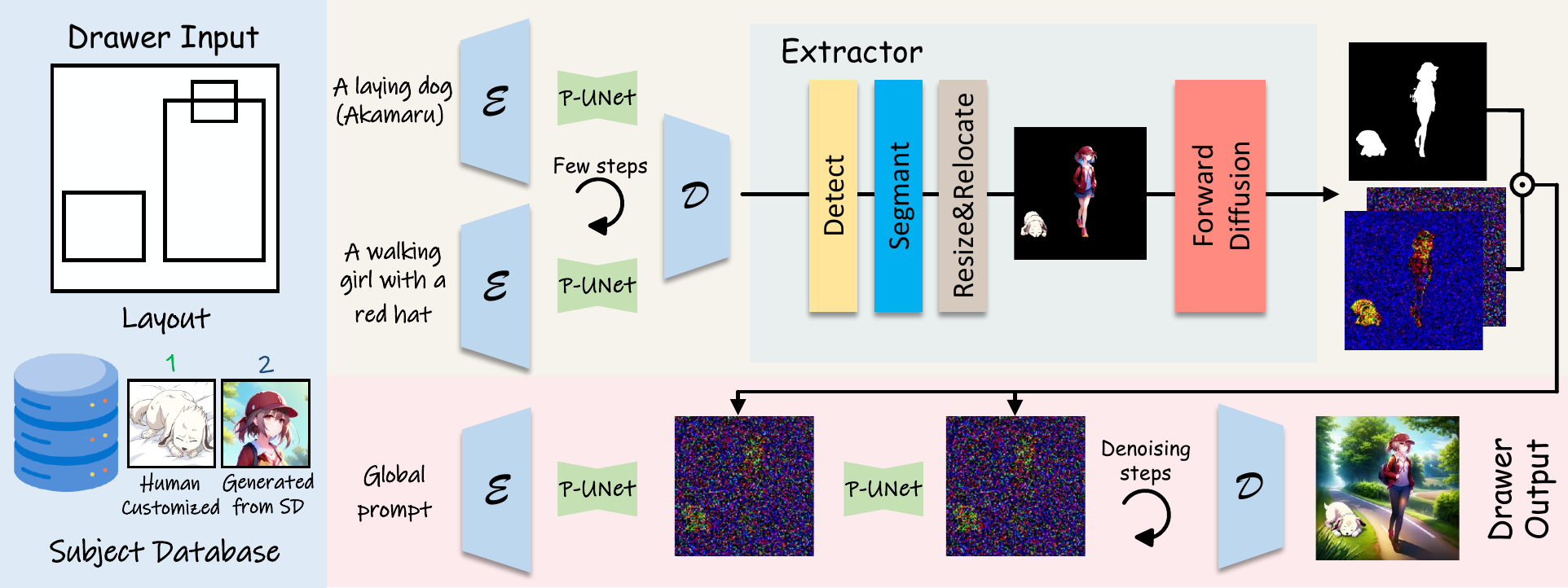}
  \caption{Overall structure of our subject-initialized generation method.}
  \label{fig: arch_subject_init}
\end{figure*}

Even with well-organized layouts, existing T2I models still face challenges in generating appealing images with consistent representations of multiple subjects without fine-tuning. Especially, spatially controllable sampling methods \cite{controlnet,gu2024mix} are unsuitable due to their reliance on complex additional inputs like sketches and skeletal points. Similarly, methods based on Fourier position encoding \cite{li2023gligen,ostashev2024moa} struggle to maintain the positions of subjects with arbitrary shapes (see Figure~\ref{fig: subject initialized}). To address these challenges and better exploit subject features, we propose a subject-initialized generation method and a parallel UNet architecture (P-UNet) in our drawer.

\textbf{Subject-initialized Generation}. Given the subject database $\mathcal{D}$, this initialization method generates latent feature maps that merge all subject features from $\mathcal{D}$ spatially according to the layout $\hat{\mathcal{O}}_{Layout}^k$, as shown in Figure \ref{fig: arch_subject_init}. Particularly, to better preserve features of small subjects and components, we first resize the bounding box of each subject to ensure that its long side reaches 1024 pixels. We then utilize the SD model with P-UNet (denoted as $\hat{\mathcal{SD}}$) to generate a single image for each subject with its corresponding resized and centered bounding box $\hat{b}_i$, so that we can conduct diffusion inversions to obtain their corresponding latent features. This can be formulated as,
\begin{equation}
     \{s_i|s_i = \hat{\mathcal{SD}}(\mathbf{h}_i, \mathbf{f}_i, \hat{b}_i), i=1,...,n\},
\end{equation}
where $s_i$ denotes the generated image of subject $i$. $\mathbf{h}_i$ is the image embedding of subject $i$ retrieved from $\mathcal{D}$, which is obtained by encoding the initially generated subject image or a user-provided reference image. We employ the pre-trained CLIP image encoder \cite{Clip} followed by the projection module of IP-Adapter \cite{ipadapter} to conduct image encoding in this paper. $\mathbf{f}_i$ denotes the text embedding of the captions of subject $i$ along with the global caption and background caption in $\hat{\mathcal{O}}_{Layout}^k$, which is also obtained via the pre-trained CLIP text encoder. 

Note that we use $s_i$ only for initialization and hence it is unnecessary to conduct the whole denoising process of $\hat{\mathcal{SD}}$ to generate a fine-grained $s_i$. Experimentally, we notice that about 1/10 of the total diffusion time steps are sufficient to generate $s_i$ for effective guidance. This strategy helps us to address the expensive extra time consumption for single-subject image generation in \cite{cheng2024theatergen}. 

To consolidate all single-subject images into one that is coherent with $\hat{\mathcal{O}}_{Layout}^k$, we utilize an extractor that consists of an open-vocabulary detection model \cite{GroundingDino} and a segmentation model \cite{SAM}. We then resize all segmented subjects and incorporate them based on their corresponding original bounding boxes into a blank guidance image $I_G$. By applying the forward diffusion process of $\hat{\mathcal{SD}}$ on ${I}_{G}$, we can project ${I}_{G}$ into the latent space of $\hat{\mathcal{SD}}$ and obtain a guidance set $\mathcal{G}$ as follows:
\begin{equation}
    \mathcal{G} = \{{\mathbf{G}}_t | {\mathbf{G}}_t = \hat{\mathcal{FD}}(\mathcal{I}_{g}, t), t=0,...,T-1\},
\end{equation}
where $\hat{\mathcal{FD}}$ denote the forward diffusion process with $T$ steps and ${\mathbf{G}}_t$ represents the multi-subject guidance at the $t$-th step. We incorporate ${\mathbf{G}}_t$ into the denoising process of $\hat{\mathcal{SD}}$ to generate our target image $I_k$ as follows:
\begin{align}
\boldsymbol{\hat{Z}}_t = 
\begin{cases}
    \boldsymbol{Z}_t \odot (1-M) + {\mathbf{G}}_t \odot M, & \text{if } t \geq rT, \\
    \boldsymbol{Z}_t,  & \text{otherwise}. \\
\end{cases}
\end{align}
Here $\boldsymbol{Z}_t$ is the latent representation of $I_k$ at time step $t$. $\odot$ denotes the element-wise product. $M$ denotes the binary segmentation mask obtained on $I_G$. $r$ is a hyperparameter controlling the starting step of applying the multi-subject guidance. We recommend setting $r$ to 0.95 since diffusion models typically generate the overall structure of subjects in early denoising steps \cite{lian2023llm,sun2024spatial}. In this way, all generated single-subject images are from the same latent space and play a role in the process of generating the image of current turn ${I}_k$.

\begin{figure*}[!t]
  \centering
  \includegraphics[width=1\textwidth]{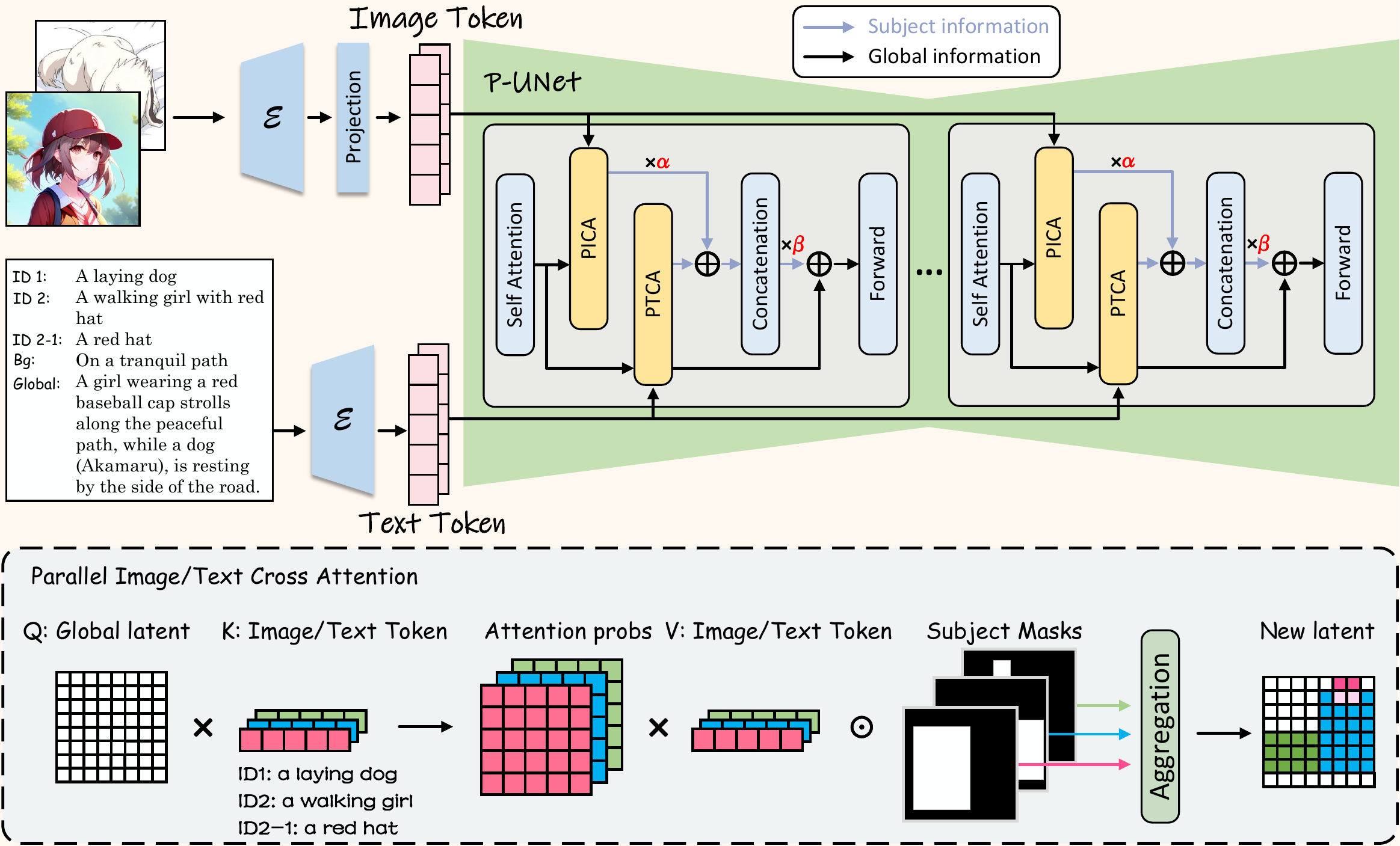}
  \caption{The overall structure of P-UNet, of which the core components are the parallel text and image cross-attention modules.}
  \label{fig: P-UNet}
\end{figure*}

\textbf{P-UNet}. The original UNet in the SD model utilizes cross-attention modules to exploit text features, which are insufficient to represent the spatial relationship and features of multiple subjects. Therefore, we propose the P-UNet that utilizes training-free layout-modulated attention modules, as shown in Figure \ref{fig: P-UNet}. With a slight abuse of notation, we still denote the input latent feature of an arbitrary UNet layer in the denoising process as $\boldsymbol{Z}$. We disentangle the original cross-attention module of the UNet layer into two parallel text and image cross-attention modules (denoted as PTCA and PICA) to refine $\boldsymbol{Z}$. These two modules have the same architecture, of which the key idea is to calculate feature similarity between $\boldsymbol{Z}$ and the per-subject text/image embedding.

Specifically, take the PTCA module as an example. We calculate the weighted representation of $\boldsymbol{Z}$ regarding the text embedding of subject $i$ as follows:
\begin{equation}
\boldsymbol{Z}_{f}^i=\operatorname{Softmax}\left(\frac{\left(W_Q \cdot \boldsymbol{Z}\right)\left(W_K \cdot {\mathbf{f}}_{i}\right)^{\top}}{\sqrt{d}}\right) W_V \cdot {\mathbf{f}}_{i},
\label{eq: cross_attn}
\end{equation}
where $W_Q$, $W_K$, and $W_V$ are the linear projections weight matrices inherited from the original cross-attention module. $d$ is the dimension of feature embedding. To reduce mutual interference among different subjects, we filter $\boldsymbol{Z}_{f}^i$ with its corresponding binary mask $R_i \in \{0, 1\}^{h \times w}$, i.e., $\boldsymbol{Z}_{f}^i := \boldsymbol{Z}_{f}^i \odot R_i$. $h$ and $w$ denote the height and width of $\boldsymbol{Z}_{f}$, respectively. The text-refined latent feature of all subjects is summarized as follows:
\begin{equation}
    \boldsymbol{Z}_{f} = M_s \odot \sum\limits_{i = 1}^n {\boldsymbol{Z}_{f}^i}.
\end{equation}
Here, $M_s = (m_{i,j})_{1 \le i \le h, 1 \le j \le w}$ is a 2D weighting matrix used to adjust features in regions overlapped by multiple subjects. We define $m_{i,j}$ as follows:
\begin{equation}
m_{i,j} = \begin{cases}
(\sum\limits_{k = 1}^n R_k(i, j))^{-1}, & \text{if } \sum\limits_{k = 1}^n R_k(i, j) > 0,\\
0, & \text{otherwise}.
\end{cases}
\end{equation}

The image-enhanced latent feature $\boldsymbol{Z}_{h}$ is calculated similarly, i.e., we replace the text embedding ${\mathbf{f}}_i$ and the weight matrices in Eq. (\ref{eq: cross_attn}) with the image embedding ${\mathbf{h}}_i$ and the linear projections weight matrices from IP-Adapter. Our final refined latent feature ${\boldsymbol{Z}^{*}}$ is calculated as follows:
\begin{equation}
\boldsymbol{Z}^{*} = \alpha \cdot \boldsymbol{Z}_{g} + (1-\alpha)\cdot(\boldsymbol{Z}_{f} +  \beta \cdot \boldsymbol{Z}_{h}),
\label{eq:final_refinement}
\end{equation}
where $\boldsymbol{Z}_{g}$ denotes the latent feature enhanced by the embedding of the global caption ($c_{glb}$ in Eq. (\ref{eq: o_manager})). $\alpha$ and $\beta$ are hyperparameters controlling the weights of subject information and reference images, respectively.

% \paragraph{Multi-turn Editing} Especially for the editing task, AutoStudio compares the current bounding boxes with the previous round's bounding boxes. If the bounding box of subject $i$ is exactly the same as the previous round's and has the same caption ($b_i^k = b_i^{k-1} \And c_i^k = c_i^{k-1} $), AutoStudio does not update the region $b_i$ in $\mathcal{I}_{k-1}$ and directly apply it to $\mathcal{I}_{k}$ to achieve the editing effect in specific regions. HH: Place this in the appendix

\section{Experiments}

\subsection{Quantitative Evaluation}

We conduct a comprehensive evaluation of AutoStudio with the chosen baseline models on CMIGBench\cite{cheng2024theatergen}. The implementation details can be seen in Appendix~\ref{appn: Implementation Details}. CMIGBench is based on story generation and multi-turn editing, comprising 8000 multi-turn scripted dialogues (4000 for each task). Following TheaterGen \cite{cheng2024theatergen}, we choose the quantitative metrics average Fréchet Inception Distance (aFID) and average character-character similarity (aCCS) to evaluate contextual consistency and average text-image similarity (aTIS) to evaluate semantic consistency among subjects. The results from Table~\ref{tab: CMIGBench} demonstrate that AutoStudio outperforms previous methods in all metrics significantly. These quantitative experimental results demonstrate the advantages of our method in generating consistent images across multi-turn interactions.

\subsection{Qualitative Evaluation}

\begin{figure*}[!t]
  \centering
\includegraphics[width=1\textwidth]{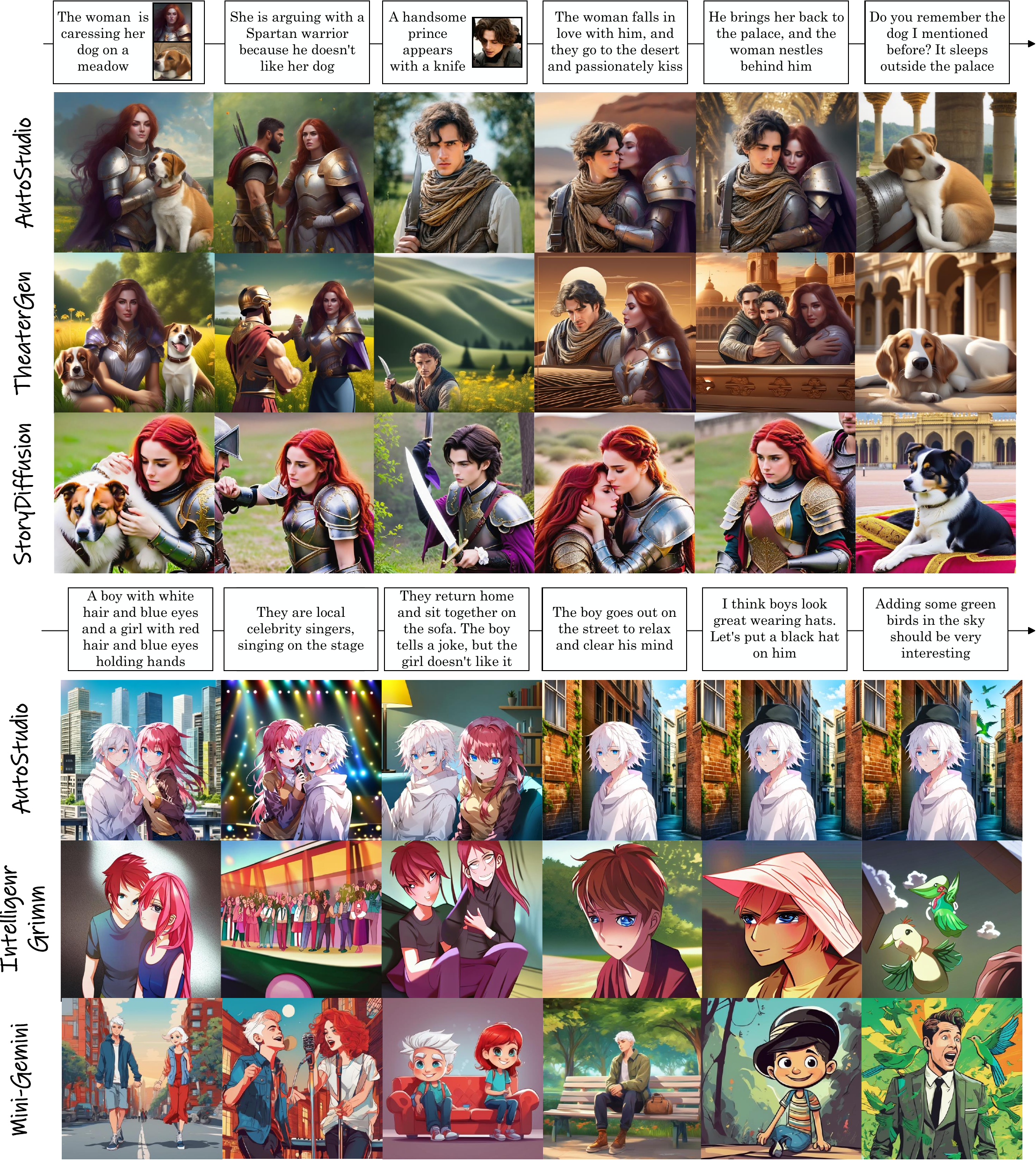}
  \caption{Visualization comparison of AutoStudio and other methods.}
  \label{fig: visualization results}
\end{figure*}

Figure~\ref{fig: visualization results} presents visualization results of multi-turn interactive image generation, showing that AutoStudio is capable of understanding natural language instructions from users and generating images with consistent subjects. Particularly, the first example of Figure~\ref{fig: visualization results} suggests that Theatergen cannot handle complex interactions between characters (such as hugging and kissing) while Mini-Gemini struggles to maintain consistent subjects. In the second example, Intelligent Grimm and StoryDiffusion fail to maintain consistency among multiple characters across multi-turn interaction and exhibit limited editing effects. More diverse generation results, such as manga book generation, can be found in Appendix~\ref{appn: More Visualization Results}.

\subsection{Ablation Study}

\textbf{Ablation on Supervisor}. We conduct an ablation study on the supervisor $\mathcal{A}_{Supervisor}$ for layout refinement. We construct a variant of our method that feeds the layout generated by $\mathcal{A}_{Layout}$ directly to $\mathcal{A}_{Drawer}$ for evaluation. The results from Table~\ref{tab: ablation} show that the performance of the baseline without $\mathcal{A}_{Supervisor}$ drops significantly. The visual examples of this ablation study are shown in Appendix~\ref{appn: ablation on layouts} as well. These results validate the effectiveness of the supervisor in layout refinement. 

\textbf{Ablation on P-UNet}. P-UNet introduces text and image information in parallel based on the layout to maintain subject consistency. To prove its effectiveness, we conducted an ablation experiment by setting the hyperparameter $\alpha$ to 1. In this way, all the information would not undergo the parallel processing operation, resulting in a significant decrease in the quantitative results, as indicated in Table~\ref{tab: ablation}. This serves as evidence for the effectiveness of P-UNet.

\textbf{Ablation on Subject-initialized Generation}. To validate the effectiveness of the subject-initialized generation method, we conduct an ablation study on CMIGBench by setting the hyperparameter $r$ to 0. The quantitative results from Table~\ref{tab: ablation} and visualization in Appendix~\ref{appn: Subject Guidance} both demonstrate that without the subject-initialized generation method, the probability of encountering subject missing and feature fusion has significantly increased.

\subsection{Human Evaluation}

We conduct a human study with 20 volunteers. In this study, each volunteer is given 10 dialogues selected from CMIGBench and each dialogue contains 12 questions (120 questions in total) on the quality of images synthesized by AutoStudio, TheaterGen, and StoryDiffusion. An example of these questions is shown in Figure~\ref{fig: wenjuan}. The results of this human study are summarized in Table~\ref{tab: human} and validate that AutoStudio is superior to existing methods in multi-turn interactive image generations.

\begin{table}[!t]
  \caption{Model performance on contextual and semantic consistency metrics.}
  \label{tab: CMIGBench}
  \centering
\resizebox{1\textwidth}{!}{
\begin{tabular}{cc|cc|cc|cc}
\toprule
\multirow{4}{*}{Diffsion version} & \multicolumn{1}{c}{\multirow{4}{*}{Model}} & \multicolumn{6}{c}{Metrics} \\
 & \multicolumn{1}{c}{} & \multicolumn{4}{c}{Contexual consistency} & \multicolumn{2}{c}{Sementic consistency}  \\
\cmidrule(lr){3-6} \cmidrule(lr){7-8}
 & \multicolumn{1}{c}{} & \multicolumn{2}{c}{aFID$\downarrow$} & \multicolumn{2}{c}{aCCS(\%)$\uparrow$}  & \multicolumn{2}{c}{aTIS(\%)$\uparrow$} \\
& & Story & Editing & Story & Editing & Story & Editing \\
\midrule
\multirow{6}{*}{SD1.5} & Mini DALL·E 3~\cite{Mini-DALLE3} & 451.59&443.71 & 54.11&52.51  & 29.81&28.13  \\
 & MiniGPT-5~\cite{MiniGPT5} & 528.3&480.7 & 43.09&44.08 & 24.93&22.95 \\
 & SEED-LLaMA~\cite{SEED} & 316&357.55 & 64.78&59.83 & 26.41&25.18 \\
 & Intelligent Grimm~\cite{liu2023intelligent} &416.29 & 464.93 & 43.73 & 48.86 &23.68 &24.63\\
 & TheaterGen~\cite{cheng2024theatergen} &  252.31 & 240.32 & 78&84.31 & 31.52&29.67 \\
 & \textbf{AutoStudio (Ours)} & \textbf{217.86} & \textbf{233.80} & \textbf{80.21} & \textbf{85.39} & \textbf{33.12} & \textbf{30.47}\\
 \midrule
\multirow{4}{*}{SDXL} & Mini DALL·E 3 & 286.21 &402.21 & 67.59& 54.40  & 32.77& 29.86   \\
 & StoryDiffusion~\cite{team2023gemini} & 253.28 & 320.25 & 80.23 & 72.03 &  39.42 & 35.19  \\
 & TheaterGen & 209.45&222.56 & 81.05&93.52  & 38.91&37.72    \\% \midrule
  & \textbf{AutoStudio (Ours)} & \textbf{196.99} & \textbf{218.44} & \textbf{84.66} & \textbf{93.54} & \textbf{39.76} & \textbf{40.02}  \\
 \bottomrule
\end{tabular}}

\end{table}

\section{Conclusion}

This paper presents AutoStudio, a novel training-free multi-agent framework that addresses multi-turn interactive image generation successfully. AutoStudio employs three LLM-based agents to interpret human intentions and generate appropriate layout guidance for SD models. Furthermore, a novel P-UNet architecture and a subject-initialized generation method are introduced to augment SD models with subject-aware features, which eventually helps to generate high-quality images with multi-subject consistency. Extensive experiments validate the superior performance of AutoStudio across various tasks, opening up new possibilities for advanced and user-friendly T2I applications.

\newpage
\appendix

\section{Appendix/ Supplemental material}
The outline of the Appendix is as follows:
\begin{itemize}
    \item Details of proposed AutoStudio;
        \begin{itemize}
            \item Details of subject-initialized generation method;
            \item Prompts for agents;
        \end{itemize}
    \item Experiment Details;
        \begin{itemize}
            \item Model implementation details;
            \item Ablation study and human evalution results;
        \end{itemize}
    \item More visualization;
        \begin{itemize}
            \item More multi-turn interactive image generation results;
        \end{itemize}
    \item Limitations and Social Impacts.
\end{itemize}

\section{Details of AutoStudio}

\subsection{Subject-initialized Generation Method}
\label{appn: Architecture of P-UNet}
The subject-initialized generation method is shown in Figure~\ref{fig: arch_subject_init}. This method effectively addresses the problem of feature loss and fusion in Multi-feature binding, as demonstrated in Figure~\ref{fig: subject initialized}. Especially for the editing instruction, AutoStudio utilizes Inpainting techniques to achieve editing effects by updating the latent representation of only the parts that require modification.

\subsection{Details of Agents}
\label{appn: Details of Agents}

Detailed conceptual descriptions and high-quality, thorough examples significantly enhance the output quality of the agent~\cite{wei2022chain,lian2023llm}. Hence we have designed task-specific prompts for each agent to facilitate their specific functions. The prompts for the subject manager, layout generator, and supervisor are as follows:

\begin{tcolorbox}[breakable, title=Subject Manager]
For the input of a story description, add detailed descriptions of fine-grained entities and output the structured description of the story:\\
    \hspace*{0.5cm} \textbf{(1) What are fine-grained entities?}\\
        \hspace*{1cm} Fine-grained entities are important components visible from the camera's perspective, including organ composition and external object composition. For humans, organ composition includes the head, body, etc.; external object composition includes clothing, accessories, etc. (Do not consider sound, emotions, or facial expressions as entities; they are only parts of entities.)\\
    \hspace*{0.5cm} \textbf{(2) How to supplement fine-grained entities?}\\
        \hspace*{1cm} (2.1) First, you need to think with an \textbf{entity-oriented mindset}: Fine-grained objects should be written in the same form and listed alongside the original object.\\
        \hspace*{1cm} (2.2) Secondly, you need to consider the descriptive naming of entities. \textbf{Entities should always be described as nouns. }For example, 'lithe body' is correct, 'body lithe' is incorrect!\\
        \hspace*{1cm} (2.3) Then, \textbf{fine-grained entities can be recursively considered but listed separately in descriptions}. For example, when considering the object 'princess,' the first layer of consideration includes gentle facial features, beautiful golden ceremonial gown, etc. The second layer considers features like delicate eyebrows starting from the face and various accessories starting from the gown.\\
        \hspace*{1cm} (2.4) Finally, provide a list of expanded entities' \textbf{[description, 'id']} line by line, without providing redundant descriptions.\\
    \hspace*{0.5cm} \textbf{(3) How to describe fine-grained entities?}\\
        \hspace*{1cm} (3.1) Descriptions of fine-grained entities consist of three parts, separated by commas: specific \textbf{"naming description, accompanying attribute description, detail description."}\\
        \hspace*{1cm} (3.2) For example, for the princess's ceremonial gown, first consider the naming description: golden ceremonial gown (including basic attributes like color and style); then consider the accompanying attribute description: princess's golden ceremonial gown (simple description plus the entity's subject); finally, consider the detail description: the satin ribbons of the golden ceremonial gown sway in the wind (showing its associated entity and movement).\\
        \hspace*{1cm} (3.3) Completeness of fine-grained entities: The ultimate goal of expanding fine-grained entities is to \textbf{fill the main body of the object in space}, so the entity must include the main part of the original object. For example, for the object 'princess,' if the long gown occupies a large space in the scene, it should be included; for animal entities, the torso occupies a large space and should be included.\\
        \hspace*{1cm} (3.4) Consideration of the number of fine-grained entities: One \textbf{main entity should have 3 to 7 fine-grained entities.} It shouldn't be too few or too many.\\
        \hspace*{1cm} (3.5) \textbf{Consistency Consideration} for Fine-Grained Entities: If the provided story is not "turn 1," there will generally be some context information included. You need to ensure that the fine-grained entities generated in this turn correspond to the IDs of the fine-grained entities in the previous context, and that the descriptions of color and features are similar.\\
        \hspace*{1cm} (3.6) \textbf{Consideration of differences} in fine-grained entities with respect to previous text: In each turn of the story, the descriptions of the main objects will differ, and the details of their fine-grained entities must conform to the main object description. Therefore, the detailed descriptions of fine-grained entities corresponding to the IDs in this turn must differ from the detailed descriptions in the previous text and cannot be completely identical.\\
    \hspace*{0.5cm} \textbf{(4) Below are two simple expansion results; you need to enrich yours further:}...\\
    \hspace*{0.5cm} \textbf{(5) This is the story you need to expand; }supplement it as richly as possible, and ensure that each main object is supplemented with fine-grained entities:\\
\begin{center}
\begin{tcolorbox}[breakable, title=The Input Format]
\begin{BVerbatim}
                <input>
                    <context>...</context>
                    <content>...</content>
                </input>
\end{BVerbatim}
\end{tcolorbox}
\end{center}
\end{tcolorbox}

\begin{tcolorbox}[breakable, title=Layout Generator]
Generate position representations of each main object and its fine-grained subsidiary entities within the given frame size for the following story:\\
    \hspace*{0.5cm} \textbf{(1) How to represent positions?}\\
    \hspace*{0.5cm} Similar to the CSS box model, a boundary box approach can well represent the position and size of objects. Taking the top-left corner of the frame as the origin: to the right is the x-axis, downward is the y-axis, with the box's width in the x direction as w, and the box's height in the y direction as h, use [x, y, w, h] to represent the position and size of any object.\\
    \hspace*{0.5cm} \textbf{(2) How to design the size of the boundary box?}\\
        \hspace*{1cm} \textbf{(2.1) Design the actual size of the main object.}\\
            \hspace*{1.5cm} (2.1.1) First, think about the actual size of the main types in the physical world: elephants are very large, mice are very small, how large, and how small?\\
            \hspace*{1.5cm} (2.1.2) Then anchor the standard size of the main types in the frame: even elephants and mice should not have too much size difference when presented in the frame. For aesthetic purposes: the maximum area of the main object should not exceed half of the frame area, the minimum area should not be less than 1/25 of the frame; the minimum area should not be less than 1/6 of the maximum area, and the area of an object like an adult generally accounts for 1/5 of the frame. For example, when the frame is [1024, 1024], the maximum area is 512*512, and the minimum area is 204*204.\\
            \hspace*{1.5cm} (2.1.3) Finally, adjust the standard size based on the description of the main object, for instance, a child is smaller than an adult, a girl is smaller than a boy, etc. The area difference caused by the description must be within 40%.\\
            \hspace*{1.5cm} (2.1.4) The size of the main object should be as large as possible! Unless the size contrast is particularly obvious, the minimum width and height of the main object should be greater than [250, 250].\\
        \hspace*{1cm} \textbf{(2.2) Design the actual proportion of the main object.}\\
            \hspace*{1.5cm} (2.2.1) First, think about the actual proportion of the main types in the physical world: people are generally tall and not wide, while boxes have relatively close width and height.\\
            \hspace*{1.5cm} (2.2.2) Then anchor the standard proportion of the main types in the frame: the proportion in the frame is closer to a square than in reality because rounder objects are cuter. For aesthetic purposes: the width and height cannot exceed twice the other side; usually, the real proportion should be slightly adjusted towards a square, with an adjustment range of less than 30%, i.e., not turning a rectangle into a square.\\
            \hspace*{1.5cm} (2.2.3) The width and height of an adult are generally [350, 600], and the width and height of a dog are [320, 300].\\
        \hspace*{1cm} \textbf{(2.3) Consider the distance, orientation, and posture of the main object,} adjusting the width and height according to the principles of closer objects being larger, side views being narrower, and squatting positions being shorter, with an adjustment range not exceeding 30\%.\\
        \hspace*{1cm} \textbf{(2.4) Design the proportion of the fine-grained subsidiary entities to the main object.}\\
            \hspace*{1.5cm} (2.4.1) For people, the head-to-body ratio is three to seven; facial features and hair are equivalent to the head and can be allocated according to the head's width and height; connected long clothing is equivalent to the body and can be allocated according to the body's width and height; short clothing can be allocated according to the entire body width and half the body height.\\
            \hspace*{1.5cm} (2.4.2) For animals, the height of horizontally arranged fine-grained entities is generally designed to be 90\%-100\% of the main object's height, with the width designed according to the attributes of the fine-grained entity, generally not less than 20\% of the main object's width. For vertical arrangement, the width should be 90\%-100\% while the height should be not less than 20\% of the main object. For example, a dog with a width and height of [320, 300] has fine-grained entities horizontally arranged, the height of the fine-grained entities should be 270, and the width should not be less than 64.\\
            \hspace*{1.5cm} (2.4.3) For objects, nested arrangement of the layout may occur: for instance, overall vertical arrangement with horizontal arrangement of entities below it. The nested arrangement should follow the above rules.\\
    \hspace*{0.5cm} \textbf{(3) How to design the position of the boundary box?}\\
        \hspace*{1cm} \textbf{(3.1) Scene coordination considerations}: set the absolute position of the boundary box reasonably according to the scene, e.g., birds flying in the air in the forest, and birds pecking on the ground on the street.\\
        \hspace*{1cm} \textbf{(3.2) Behavior interaction considerations}: determine the relative position of the boundary box based on the interactions between different main objects, e.g., people hugging are closer, and people lifted up are positioned higher.\\
        \hspace*{1cm} \textbf{(3.3) Object spacing considerations}: if there is no interaction between the main objects, the spacing should be as large as possible, and the overlap between the main objects in the frame should be minimized.\\
        \hspace*{1cm} \textbf{(3.4) Composition effect considerations}: the primary goal in designing the position is to find the center of the frame, allowing the centroids of all objects to be distributed as close to the center or slightly below the center as possible. Concentrating objects on one side of the frame makes the image look ugly, so composition should refer to some typical composition methods, such as central composition, horizontal line composition, vertical line composition, and symmetrical composition.\\
        \hspace*{1cm} \textbf{(3.5) The positional structure of fine-grained subsidiary entities and the main object}:\\
            \hspace*{1.5cm} (3.5.1) First, consider whether the fine-grained subsidiary entity is inside or outside the main object. Generally, accessories like hats and crowns will be outside the main object, closely attached to it, while others are usually inside the main object.\\
            \hspace*{1.5cm} (3.5.2) Next, consider the layout of fine-grained subsidiary entities: for people, horizontal layout can be used, i.e., the head (facial features and hair can be regarded as the head) is at the top, and the body (clothing and torso can be regarded as the body) is at the bottom; for side-view animals, vertically arrange their fine-grained subsidiary entities according to their orientation, e.g., the dog's head on the left and body on the right; for complex objects or animals, both horizontal and vertical layouts can be used simultaneously: for example, a house can be vertically arranged into the roof and the body of the house, with windows and doors horizontally arranged within the body.\\
            \hspace*{1.5cm} (3.5.3) Finally, finalize the boundary box shape: the vertical axis of horizontally arranged fine-grained entities should align with the main object's vertical axis, and the horizontal axis of vertically arranged entities should also align with the main object's horizontal axis. Then, tightly fill the main object with the boundary box to complete the boundary box shape. Here is an example, with a simplified description, but the actual output should follow the input description:\\
\begin{center}
\begin{tcolorbox}[breakable, title=The Description Example]
\begin{BVerbatim}
            ['house', [0, 0, 400, 300], '1']
            ['roof', [20, 15, 360, 120], '1-1']
            ['Windows', [20, 150, 140, 135], '1-2']
            ['Gate', [180, 150, 200, 135], '1-3']
\end{BVerbatim}
\end{tcolorbox}
\end{center}

    \hspace*{0.5cm} \textbf{(4) How to combine the above content when designing the boundary box:}\\
        \hspace*{1cm} \textbf{(4.1) Format of the above content}: When inputting, the above content will be given in <context>CONTEXT</context>, which will include the output content of 0-3 previous segments.\\
        \hspace*{1cm} \textbf{(4.2) Consistency considerations of position and size with the above content}: If the given story is not turn 1, some previous information is usually provided, and you need to ensure that the position and size designed this time are consistent with the previous ones, with the same ID corresponding to similar color and feature descriptions as the previous ones.\\
        \hspace*{1cm} \textbf{(4.3) Difference considerations of position and size with the above content}: Each round of the story has different descriptions of objects, and the design of their positions and sizes must follow the descriptions of the objects. Therefore, the position and size of the corresponding IDs in this round should be different from the previous ones and cannot be identical.\\
        \hspace*{1cm} \textbf{(4.4) Principle of this round's main objects}: The output of the current round only needs to consider the IDs that exist in the current round, without paying attention to objects with IDs that existed in the previous round but do not exist in the current round.\\
    \hspace*{0.5cm} \textbf{(5) Here are two simple examples of position representations. You need to learn the following examples' position representation methods and formats and strictly design your results according to the format:}\\
        \hspace*{1cm} (5.1) Here is the input and output of the first example:...\\
        \hspace*{1cm} (5.2) Here is the input and output of the second example:...\\
    \hspace*{0.5cm} \textbf{(6) Here is the story for which you need to supplement the position representation.} Each main object and fine-grained entity needs to supplement the position representation:
\begin{center}
\begin{tcolorbox}[breakable, title=The Input Format]
\begin{BVerbatim}
                <input>
                    <size>[1024, 1024]</size>
                    <context>...</context>
                    <content>...</content>
                </input>
\end{BVerbatim}
\end{tcolorbox}
\end{center}
    \hspace*{0.5cm} \textbf{(7) Next, please output} \texttt{<output>}Yours Output\texttt{</output>}, do not output extra content:
\end{tcolorbox}

\begin{tcolorbox}[breakable, title=Supervisor]
Below is a story with added bounding boxes for object positions and sizes. You need to check whether the format and content of the story results are compliant and provide your own modification advice:\\
    \hspace*{0.5cm} \textbf{(1) Format specifications:}\\
        \hspace*{1cm} \textbf{(1.1) Check the overall structure}: Each object is given in a list of ["description", bounding box, "id"], line by line, without any extra content.\\
        \hspace*{1cm} \textbf{(1.2) Check the structure of the description part}: The description of fine-grained entities includes three parts, separated by commas: specific "naming description, attribute description, detailed description". The description of the main body has only the naming description.\\
        \hspace*{1cm} \textbf{(1.3) Check the structure of the bounding box part}: The bounding box does not need to be enclosed in "", it should be represented in list form: [x, y, w, h].\\
        \hspace*{1cm} \textbf{(1.4) Check the format of the quotes}: In the output format, both description and id should be enclosed in double quotes, not single quotes. However, when using quotes within the description, use single quotes.\\
        \hspace*{1cm} \textbf{(1.5) Example}:...\\
    \hspace*{0.5cm} \textbf{(2) Content specifications:}\\
        \hspace*{1cm} \textbf{(2.1) Check if the size of the object is designed correctly}:\\
            \hspace*{1.5cm} (2.1.1) Check the \textbf{relative size of the main body}: First, review the main bodies in the story, which one is big? Which one is small? Is the size relationship correct? Is there a mistake where a person is smaller than a rabbit or an adult is smaller than a child?\\
            \hspace*{1.5cm} (2.1.2) Check the \textbf{absolute size of the main body}: Is the size proportion of the main body to the frame correct? Is it too big? Is it too small? Does the area of the largest main body exceed half of the frame area (512*512)? Does the area of the smallest main body less than 1/25 of the frame (204*204)? Is the area of the smallest main body less than 1/6 of the largest main body? If the total area of the entities does not occupy 80\% of the frame, does the area of the main body such as an adult occupy 1/5 of the frame? Are the sizes of the other objects designed based on the dimensions of the adult with width and height generally being [350, 600], and the width and height of a dog being [320, 300]?\\
        \hspace*{1cm} \textbf{(2.2) Check} if \textbf{the proportion relationship} between the main body and the fine-grained entities is correctly designed:\\
            \hspace*{1.5cm} (2.2.1) Proportion check of \textbf{human fine-grained entitie}s: Does the proportion distribution of the human satisfy the ratio of head to body as three to seven, and the facial features and hair are equivalent to the head? They can all be allocated according to the width and height of the head. Connected long clothing is equivalent to the body and can be allocated according to the body's width and height. Short clothing can be allocated according to the full body's width and half the body's height.\\
            \hspace*{1.5cm} (2.2.2) Proportion check of \textbf{animal fine-grained entities}: The proportion distribution of animals should meet the design of horizontally arranged fine-grained entities, with height generally designed as 90\%-100\% of the main body height, and width designed according to the attributes of the fine-grained entities, generally not less than 20\% of the main body's width. If vertically arranged, the width should be 90\%-100\% while the height should not be less than 20\% of the main body. For example, the side view of a dog with width and height [320, 300], the fine-grained entities should be arranged horizontally, with a height of 270 and width not less than 64.\\
            \hspace*{1.5cm} (2.2.3) Proportion check of \textbf{object fine-grained entities}: The proportion distribution of objects should meet the possible nesting of arrangement methods of objects: for example, overall vertical arrangement, with lower entities arranged horizontally, therefore nesting follows the above rules.\\
        \hspace*{1cm} \textbf{(2.3) Check if the position of the object is designed correctly:}\\
            \hspace*{1.5cm} (2.3.1) \textbf{Scene coordination consideration}: Set the absolute position of the bounding box reasonably according to the scene, for example, birds flying in the air in a forest, birds might be pecking at the ground on a street.\\
            \hspace*{1.5cm} (2.3.2) \textbf{Behavioral interaction consideration}: Determine the relative position of the bounding boxes based on the interaction between different main body objects, such as people hugging being closer together, people being lifted higher.\\
            \hspace*{1.5cm} (2.3.3) \textbf{Object interval consideration}: If there is no necessary interaction or the frame is already occupied, the interval between objects needs to be as large as possible!!! Overlapping of main body objects in the frame should be minimized!!! The standard is no overlap if no interaction, and when overlap is necessary, the overlap between main bodies should not exceed 25\%!\\
            \hspace*{1.5cm} (2.3.4) \textbf{Composition effect consideration}: The primary goal of designing positions is to find the center of the frame, allowing the total mass center of all objects to be distributed around the center or slightly below the center as much as possible. Concentrating objects on one side of the frame will make the picture look bad, so composition needs to refer to some typical composition methods such as central composition, horizontal line composition, vertical line composition, symmetrical composition methods, etc.\\
        \hspace*{1cm} \textbf{(2.4) Check} if \textbf{the position relationship} between the main body and the fine-grained entities is designed according to the specifications: How are the fine-grained entities arranged within the main body? Horizontally, vertically, or otherwise? Here are the specific specifications:\\
            \hspace*{1.5cm} (2.5.1) First, consider whether the attached fine-grained entities are inside or outside the main body. Generally, accessories like hats and crowns will be outside the main body, close to it, while others are generally inside the main body.\\
            \hspace*{1.5cm} (2.5.2) Second, consider the arrangement of the attached fine-grained entities: For humans, it can be arranged horizontally, with the head (facial features, hair can be considered as the head) on top and the body (clothing, body can be considered as the body) below. For side-view animals, arrange the attached fine-grained entities vertically according to their orientation, such as the dog's head on the left and body on the right. Complex objects or animals can use both horizontal and vertical arrangements simultaneously: for example, a house can be vertically arranged as the roof and body, with the body horizontally arranged as windows-doors, etc.\\
    \hspace*{0.5cm} \textbf{(3) Your task:}\\
        \hspace*{1cm} \textbf{(3.1) If there are formatting errors, correct the formatting errors} (the formatting errors here refer to the structured format mentioned in (1): ["naming description, attribute description, detail description", [x, y, w, h], "id"]).\\
        \hspace*{1cm} \textbf{(3.2) Check the overlap ratio between main bodies.} If it exceeds 30\%, re-layout the frame (note that the bounding box of the main body should not exceed the frame).\\
        \hspace*{1cm} \textbf{(3.3)} If there are \textbf{errors in the size of the objects}, redesign them (the size of the objects has been exaggerated, and should not be smaller than 1/25 of the frame. The objects should be as large as possible without affecting relative size and overlap).\\
        \hspace*{1cm} \textbf{(3.4)} If there are \textbf{proportion relationship errors} between the main body and fine-grained entities, redesign.\\
        \hspace*{1cm} \textbf{(3.5)} If there are \textbf{position errors} of objects, redesign.\\
    \hspace*{0.5cm} \textbf{(4) Object interval is often an issue that needs adjustment.} Here is a modification example, you need to refer to the output to see what modifications were made to the input and provide your own modification plan after the input:...\\
    \hspace*{0.5cm} \textbf{(5) Below is the input content:}...\\

\begin{center}
\begin{tcolorbox}[breakable, title=The Input Format]
\begin{BVerbatim}
                <input>
                    <size>[1024, 1024]</size>
                    <content></content>
                </input>
\end{BVerbatim}
\end{tcolorbox}
\end{center}

    \hspace*{0.5cm} \textbf{(6)} \textbf{Generally speaking, the story layout is not compliant, so the subject should be as large as possible}. If there is no overlap between subjects, do not make the subjects smaller!) Next, please output your result formatted by the below format and do not output extra content.\\

\begin{center}
\begin{tcolorbox}[breakable, title=The Output Format]
\begin{BVerbatim}
                <output>
                    <advice>...</advice>
                </output>
\end{BVerbatim}
\end{tcolorbox}
\end{center}
\end{tcolorbox}

\section{Experiment Details}

\subsection{Implementation Details}
\label{appn: Implementation Details}

Our AutoStudio is a training-free and versatile framework that is compatible with most existing LLM architectures and diffusion models. In our experiments, we choose GPT-4o~\cite{gpt4o} to implement $\mathcal{A}_{Manager}$, $\mathcal{A}_{Layout}$, and $\mathcal{A}_{Supervisor}$, while SD1.5/SDXL for 
$\mathcal{A}_{Drawer}$. We adopt the DDIM sampler with 30 steps in $\mathcal{A}_{Drawer}$. The subject guidance factor $r$ is set to 0.95. The parallel introduction factor $\alpha$ and the image intensity factor $\beta$ in Eq. (\ref{eq:final_refinement}) are set to 0.2 and 0.7, respectively. The evaluation process on CMIGBench is completed over a period of 60 GPU hours, utilizing one NVIDIA GeForce RTX 3090 GPU with 25GB of memory.

In addition to the results reported in Theatergen, we also compare the recent models in CMIGBench. The deployment details of these models are as follows.

\paragraph{StoryDiffusion} introduces Consistent Self-Attention to maintain subject consistency in a generation batch. However, this method does not support natural language input nor on-the-fly interaction. To compare the effectiveness, we provid on-the-fly dialogue as a one-time input. The model variant evaluated in our experiments is "StoryDiffusion Version 0.01". The evaluation process on CMIGBench is completed over a period of 50 GPU hours, utilizing one NVIDIA GeForce RTX 3090 GPU with 25GB of memory.

\paragraph{Intelligent Grimm} utilizes a visual language context module that can generate the current frame by adjusting for the corresponding textual prompts and preceding image-caption pairs. The model is based on SD1.5. The evaluation process on CMIGBench is completed over a period of 35 GPU hours, utilizing one NVIDIA GeForce RTX 3090 GPU with 25GB of memory.

\subsection{Ablation Study Results}
The results of the ablation study on CMIGBench benchmark are shown in Table~\ref{tab: ablation}. The effectiveness of the proposed AutoStudio is demonstrated by the fact that the absence of any component results in a decrease in all metrics. This clearly indicates that every component of AutoStudio plays a crucial role in enhancing performance.

\begin{table}[!h]
  \caption{Results of ablation study.}
  \label{tab: ablation}
  \centering
\resizebox{1\textwidth}{!}{
\begin{tabular}{cl|cc|cc|cc}
\toprule
\multirow{4}{*}{Diffsion version} & \multicolumn{1}{c}{\multirow{4}{*}{Model}} & \multicolumn{6}{c}{Metrics} \\
 & \multicolumn{1}{c}{} & \multicolumn{4}{c}{Contexual consistency} & \multicolumn{2}{c}{Sementic consistency}  \\
\cmidrule(lr){3-6} \cmidrule(lr){7-8}
 & \multicolumn{1}{c}{} & \multicolumn{2}{c}{aFID$\downarrow$} & \multicolumn{2}{c}{aCCS(\%)$\uparrow$}  & \multicolumn{2}{c}{aTIS(\%)$\uparrow$} \\
& & Story & Editing & Story & Editing & Story & Editing \\
\midrule
\multirow{4}{*}{SD1.5} 
 & \textit{w/o Supervisor} & 390.63 & 283.38 &53.09 &  69.15 & 29.14 & 27.22\\
 & \textit{w/o P-UNet} &453.08 & 472.63 &52.46 & 54.40 &26.86 & 17.89\\
 & \textit{w/o Subject guidance} &277.41 & 283.55 &65.96 & 69.08 &29.26 & 27.23 \\
 & \textbf{AutoStudio (Ours)} & \textbf{217.86} & \textbf{233.8} & \textbf{80.21} & \textbf{85.39} & \textbf{33.12} & \textbf{30.47}\\
 \bottomrule
\end{tabular}}
\end{table}

\subsection{Human Evaluation Results}
\label{appn: Human Evaluation}
\begin{figure*}[!h]
  \centering
  \includegraphics[width=1\textwidth]{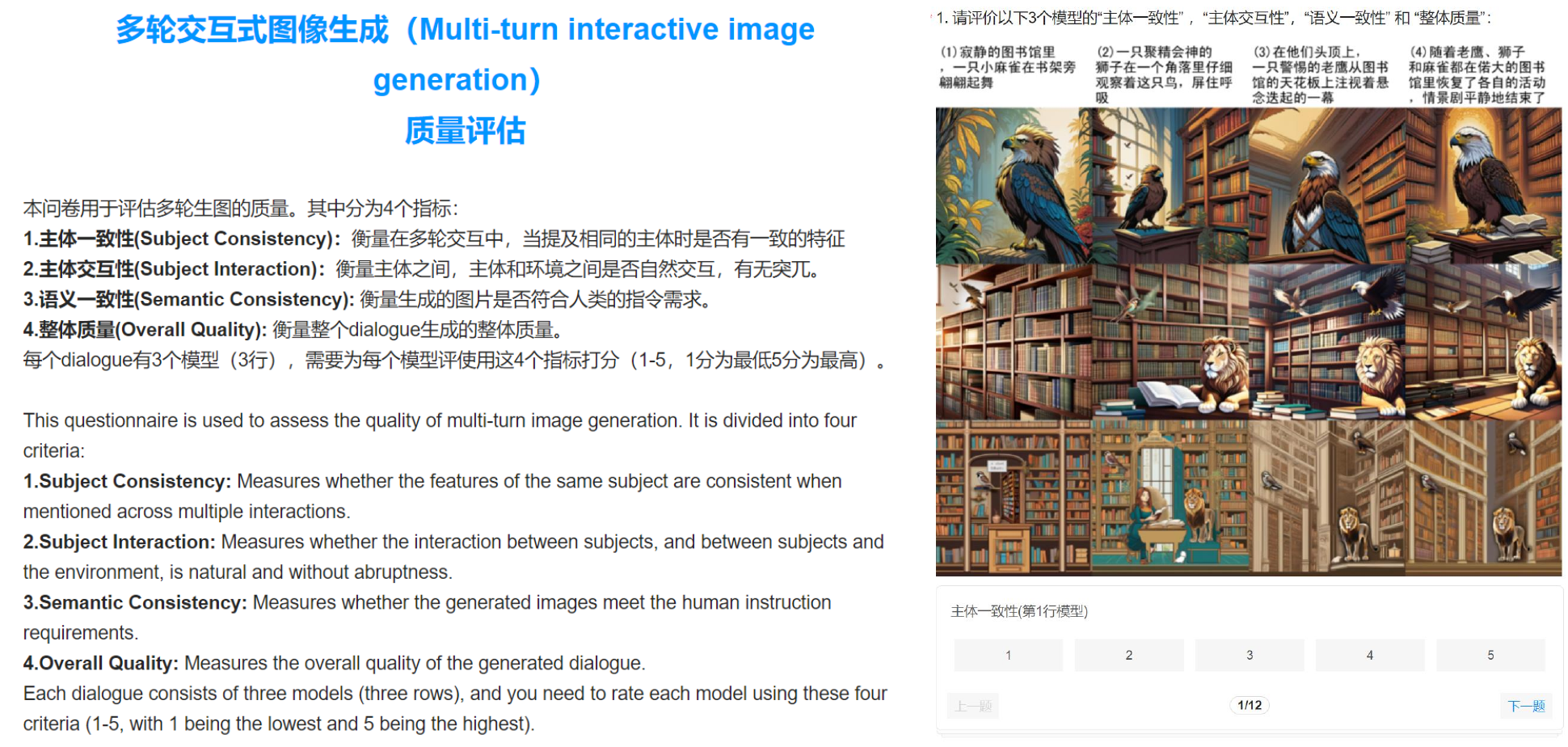}
  \caption{Screenshot of the questionnaire for human evaluation.}
  \label{fig: wenjuan}
\end{figure*}

We conducted a human evaluation for AutoStudio using a questionnaire, as depicted in Figure~\ref{fig: wenjuan}. The results of the evaluation, presented in Table~\ref{tab: human}, clearly demonstrate that AutoStudio surpasses other models in all four key metrics: Subject Consistency, Subject Interaction, Semantic Consistency, and Overall Quality. This indicates that AutoStudio consistently performs better than its counterparts across various aspects, reaffirming its superiority.

\begin{table}[!h]
  \caption{Results of human evaluation.}
  \label{tab: human}
  \centering
\resizebox{1\textwidth}{!}{
\begin{tabular}{lcccc}
\toprule
 & Subject Consistency & Subject Interaction & Semantic Consistency & Overall Quality \\
\midrule
StoryDiffusion & 2.33 & 3.3 & 2.53 & 3.11  \\
TheaterGen & 2.60 & 1.91 & 2.56 & 2.8\\
\textbf{AutoStudio} & \textbf{3.66} & \textbf{3.8} & \textbf{3.56} & \textbf{3.93} \\
\bottomrule
\end{tabular}}
\end{table}

\section{More Visualization Results}
\label{appn: More Visualization Results}

\subsection{Visualization results on CMIGBench}
\label{appn: ablation on layouts}

\begin{figure*}[!h]
  \centering
  \includegraphics[width=0.7\textwidth]{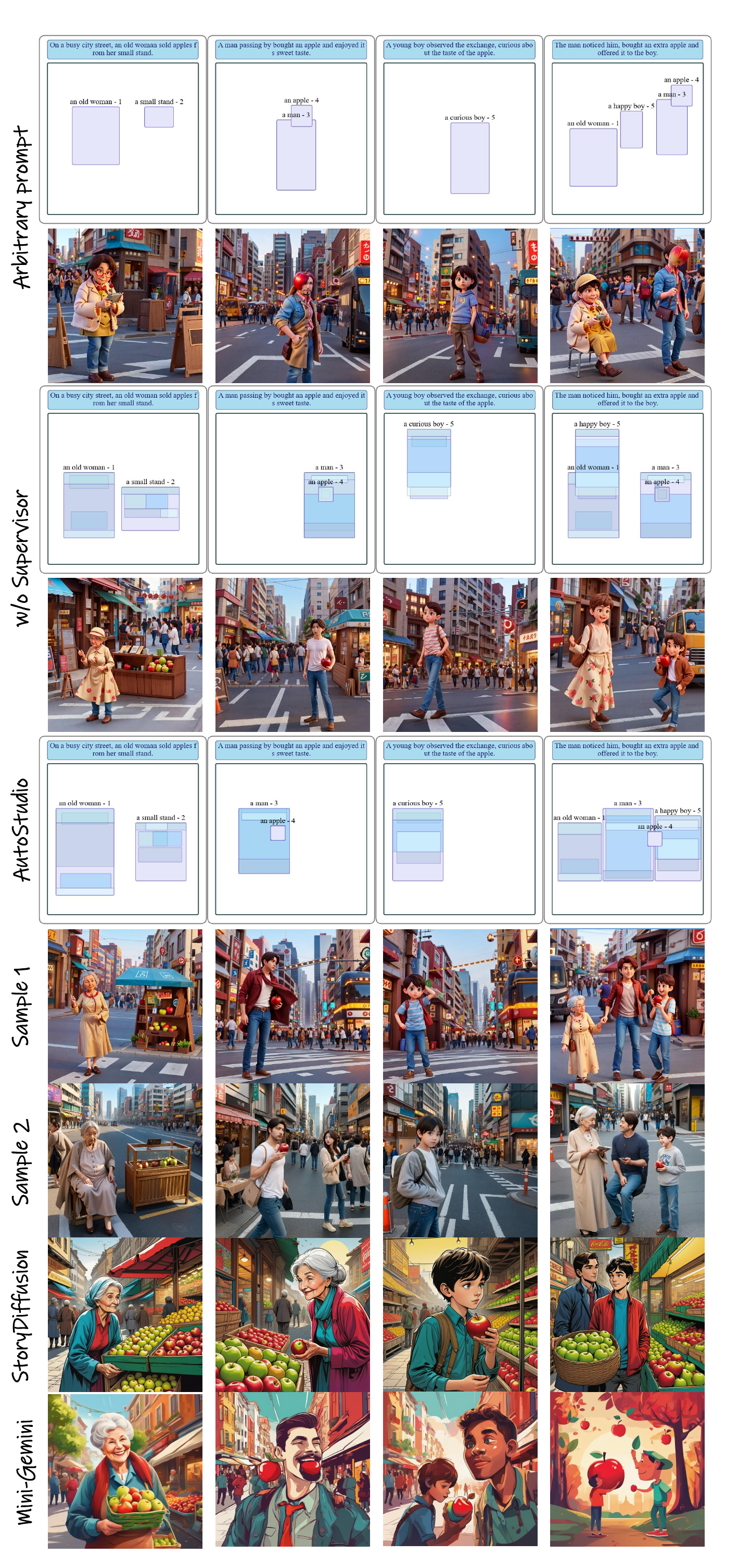}
  \caption{Visualizations Comparison on CMIGBench.}
  \label{fig: lx1}
\end{figure*}

\begin{figure*}[!h]
  \centering
  \includegraphics[width=0.7\textwidth]{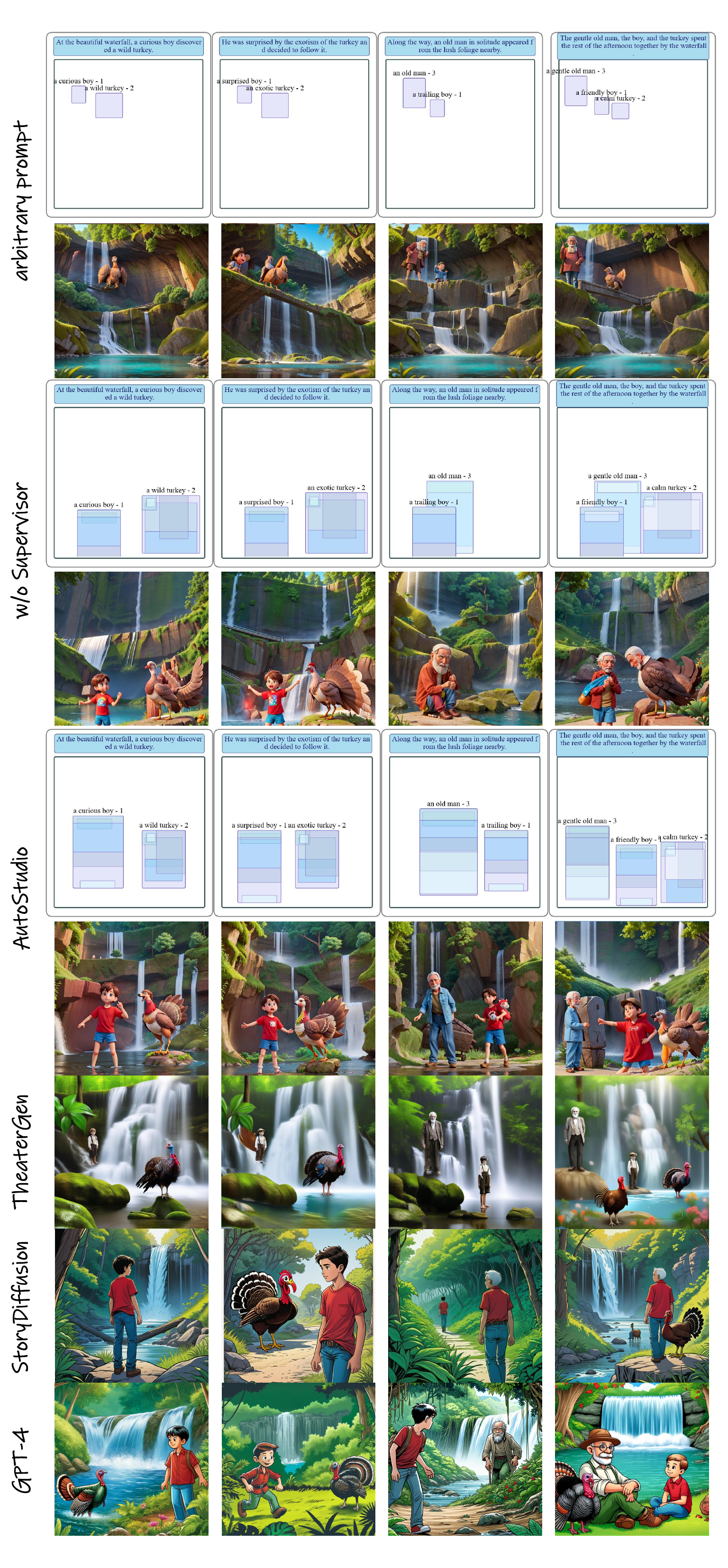}
  \caption{Visualizations Comparison on CMIGBench.}
  \label{fig: lx2}
\end{figure*}

\begin{figure*}[!h]
  \centering
  \includegraphics[width=0.8\textwidth]{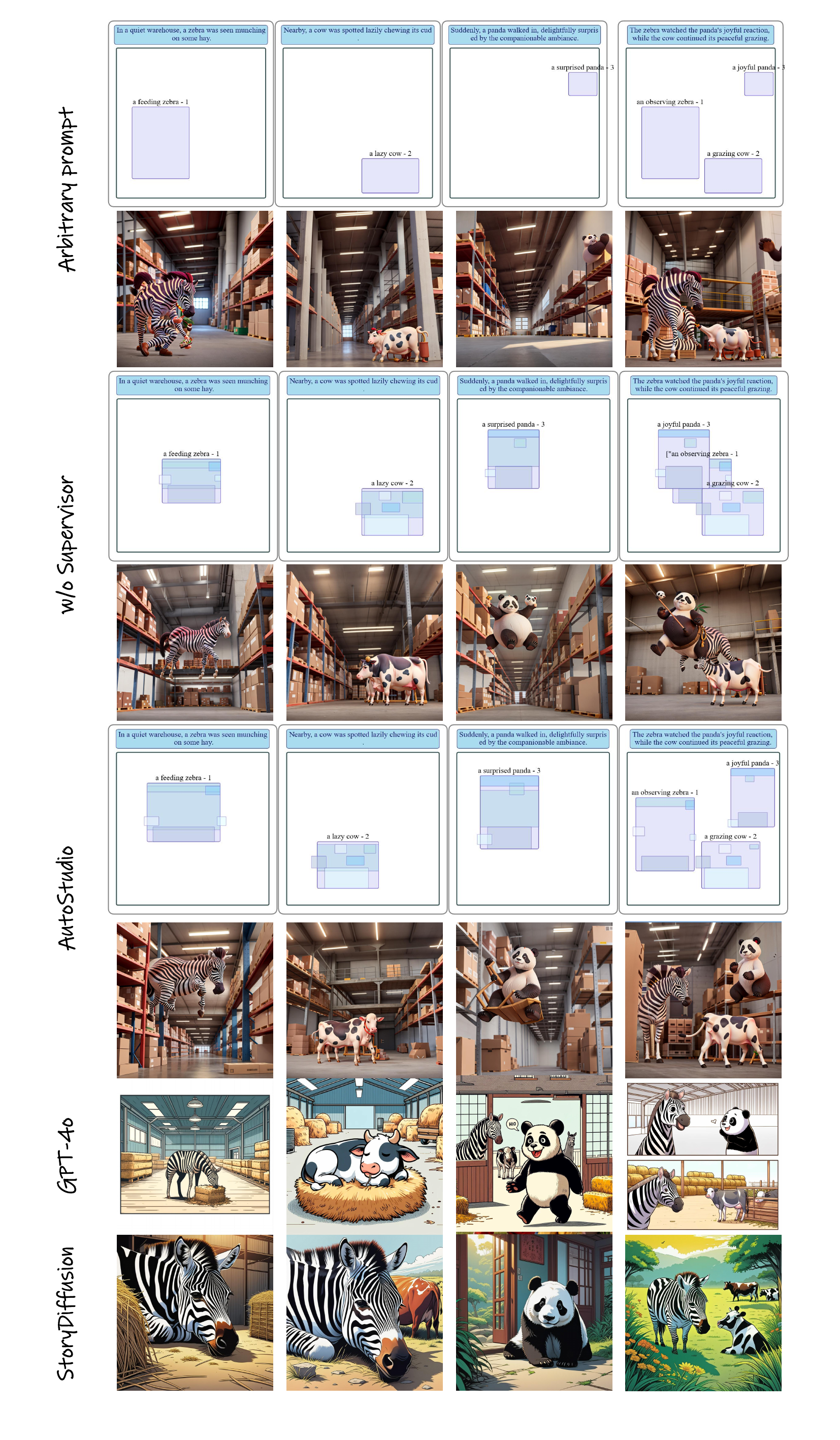}
  \caption{Visualizations Comparison on CMIGBench.}
  \label{fig: lx3}
\end{figure*}

\begin{figure*}[!h]
  \centering
  \includegraphics[width=0.7\textwidth]{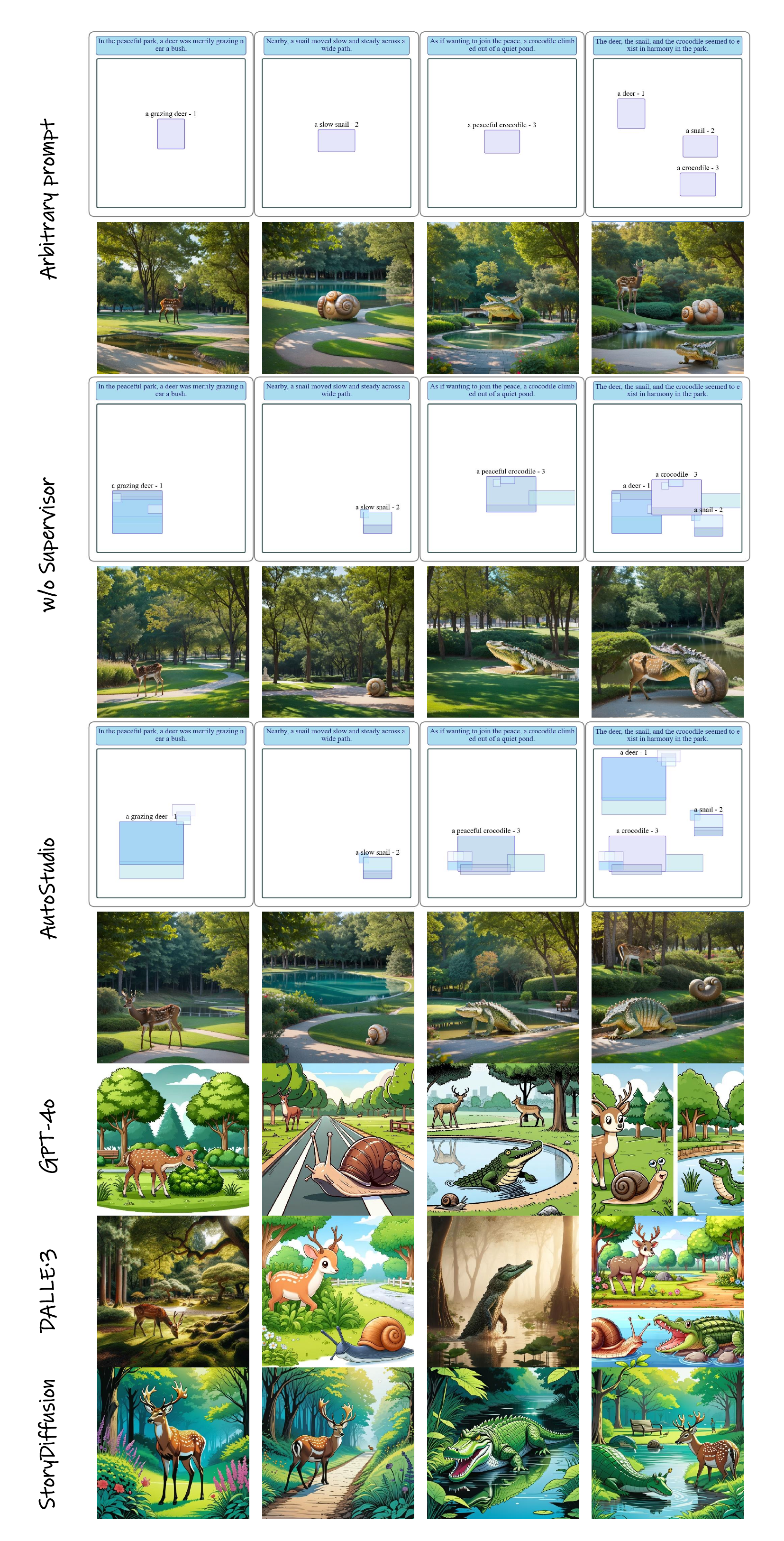}
  \caption{Visualizations Comparison on CMIGBench.}
  \label{fig: lx4}
\end{figure*}

Figure~\ref{fig: lx1} to Figure~\ref{fig: lx4} demonstrate the open-ended story generation results on CMIGBench with visualizations of layouts. A comparison is conducted between the supervisor-refined layout, the original layout, and the layout generated using a arbitrary prompt ("Generating a layout for this instruction"). The results indicate that the absence of the supervisor refinement process leads to layouts that exhibit overlapping elements, unreasonable arrangement, and incorrect sizes, such as a turkey's box being larger than a person's. Moreover, layouts generated using a arbitrary prompt show significantly inferior quality and lack fine-grained features, thus highlighting the effectiveness of the layout generator and the supervisor.

In addition, these layouts are employed to generate images, and a comparison is conducted with both open-source methods and state-of-the-art closed-source models such as GPT-4o~\cite{gpt4o} and DALLE ·3~\cite{DALLE3}. The results demonstrate that, with the supervisor, AutoStudio preserves multi-subject consistency while generating high-quality images during on-the-fly interaction with users. This underscores the capability of AutoStudio to maintain coherence and produce visually appealing images when engaged in dynamic interactions.

\subsection{Visualization of Multi-subject Image Generation}
\label{appn: Subject Guidance}
\begin{figure*}[!h]
  \centering
  \includegraphics[width=\textwidth]{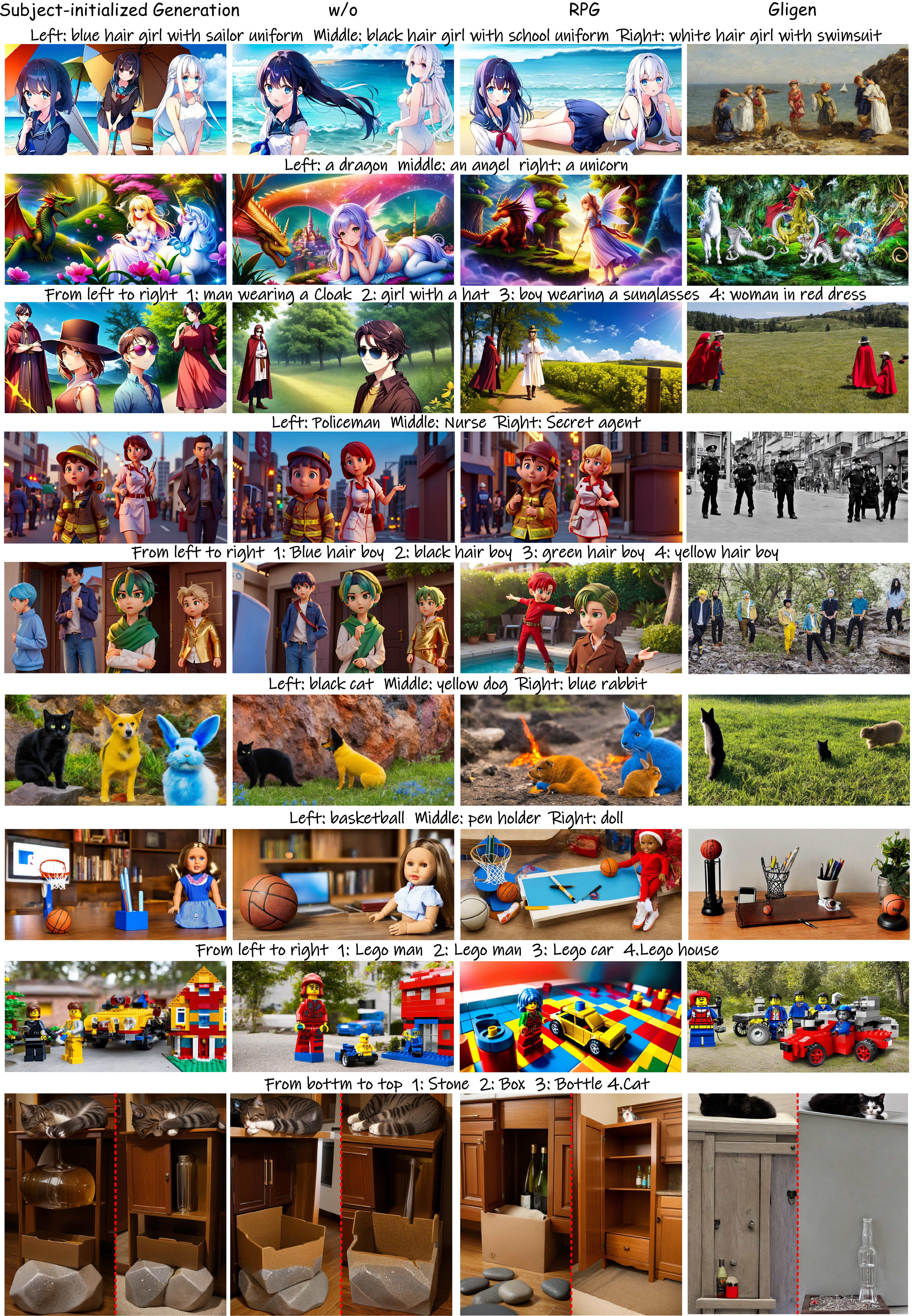}
  \caption{Visualization comparison between AutoStudio, AutoStudio w/o subject-initialized generation method, RPG and Gligen.}
  \label{fig: subject initialized}
\end{figure*}

The effectiveness of the proposed subject-initialized generation method in addressing subject missing and subject fusion issues is further validated through the visualization of multi-subject image generation. The results presented in Figure~\ref{fig: subject initialized} clearly demonstrate that in the absence of the subject-initialized generation method, the generated images often suffer from problems like missing characters or feature fusion. It is worth noting that even state-of-the-art methods such as RPG~\cite{yang2024mastering} and Gligen~\cite{li2023gligen} exhibit poor performance when generating non-square images.

\subsection{Manga Book Generation}
\begin{figure*}[!h]
  \centering
  \includegraphics[width=1\textwidth]{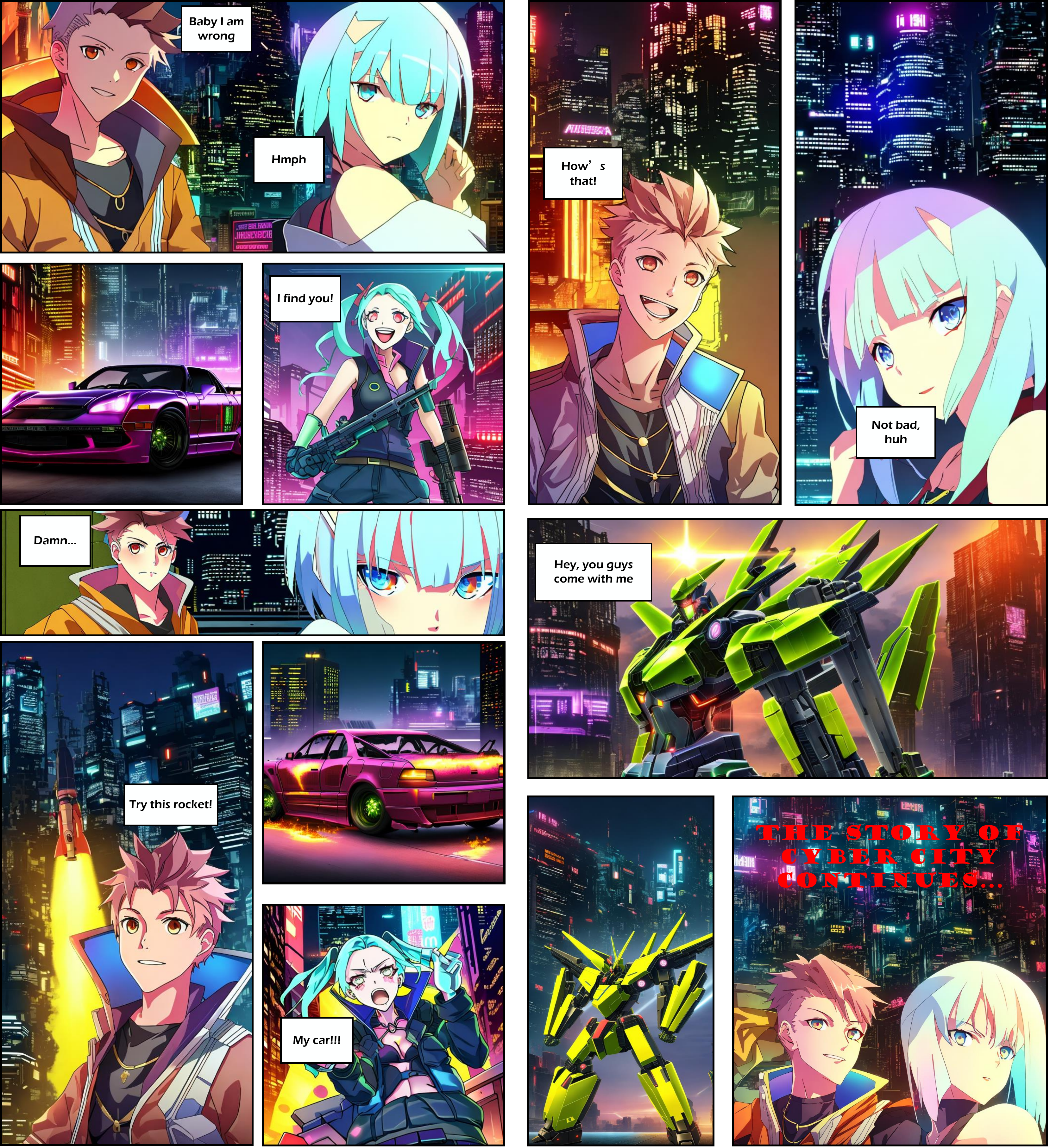}
  \caption{Manga book generation results.}
  \label{fig: mamga1}
\end{figure*}

\begin{figure*}[!h]
  \centering
  \includegraphics[width=1\textwidth]{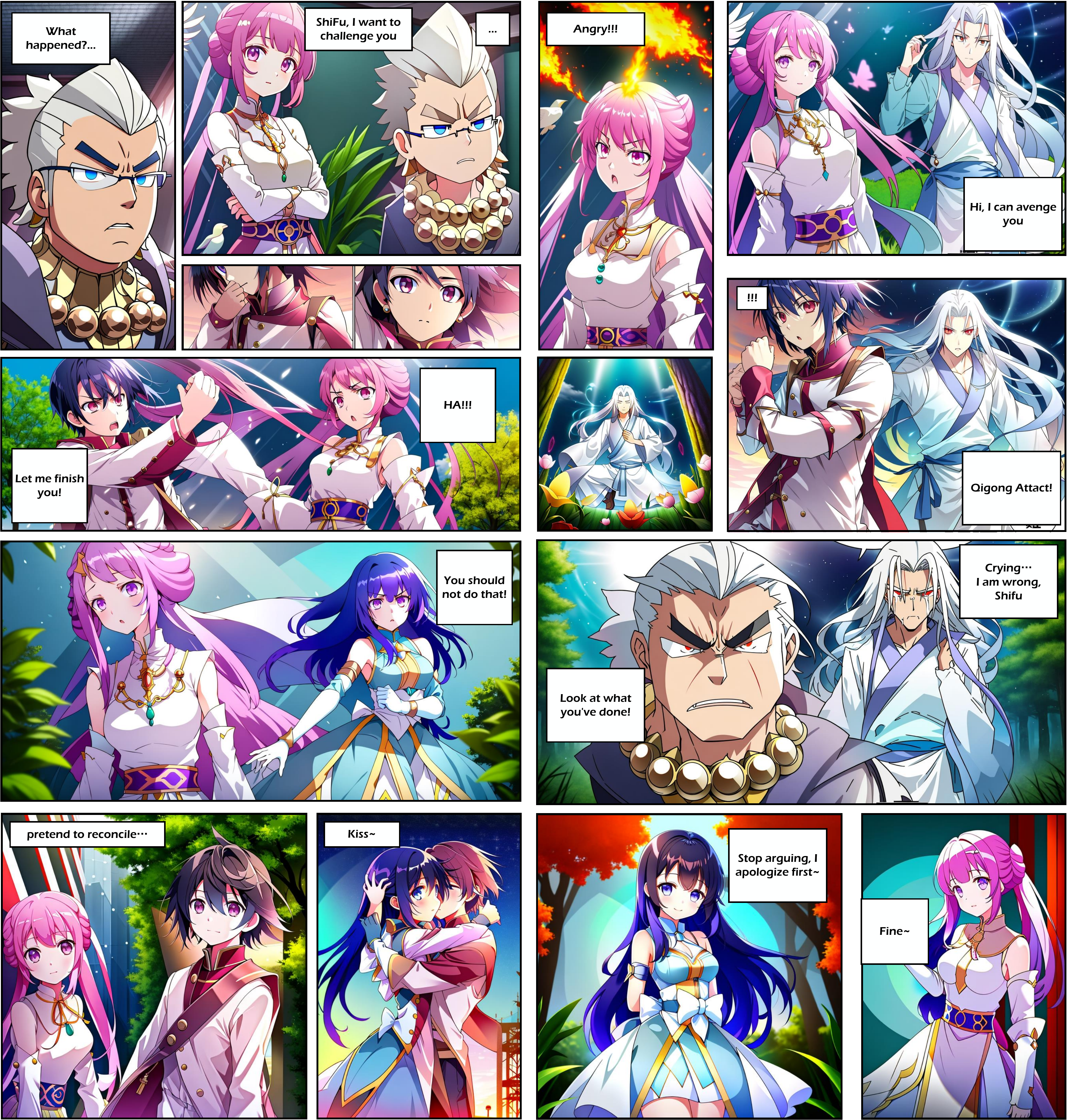}
  \caption{Manga book generation results.}
  \label{fig: mamga2}
\end{figure*}

AutoStudio provides support for arbitrary input shapes, enabling the generation of manga books through multi-turn interaction. The results depicted in Figure~\ref{fig: mamga1} and \ref{fig: mamga2} demonstrate that AutoStudio can maintain consistency among multiple characters and generate high-quality manga books with rich plotlines. AutoStudio outperforms StoryDiffusion in manga book generation due to its on-the-fly interaction capability. Unlike StoryDiffusion, AutoStudio allows for immediate adjustments and revisions to unsatisfactory images without the need to regenerate the entire manga. Furthermore, AutoStudio has the capability to generate manga images of arbitrary size, thereby significantly enhancing the visual dynamism and tension within the manga.

\subsection{Multi-turn Interactive Image Generation}

\begin{figure*}[!h]
  \centering
  \includegraphics[width=1\textwidth]{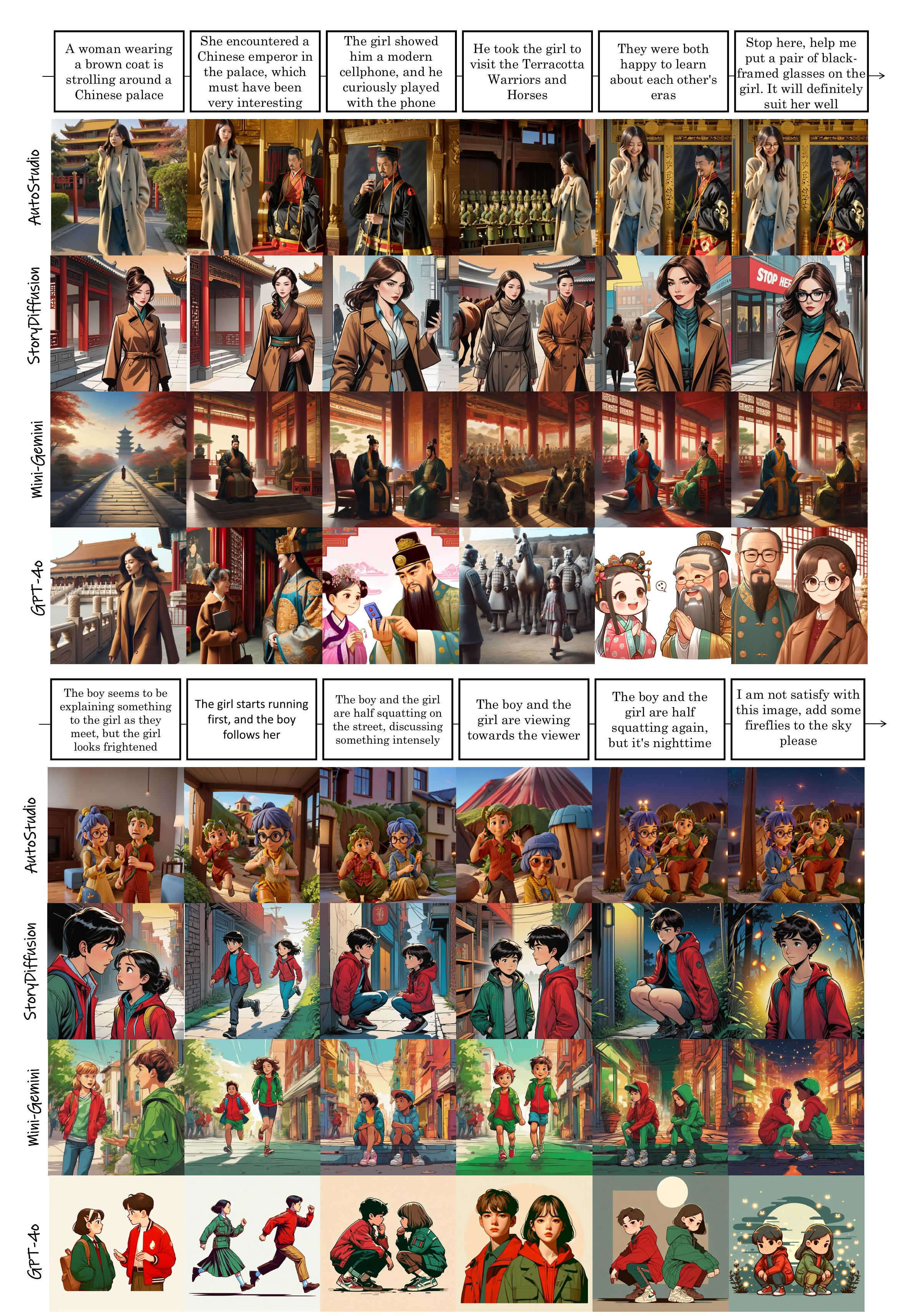}
  \caption{Multi-turn Interactive Image Generation results.}
  \label{fig: cjh1}
\end{figure*}

\begin{figure*}[!h]
  \centering
  \includegraphics[width=1\textwidth]{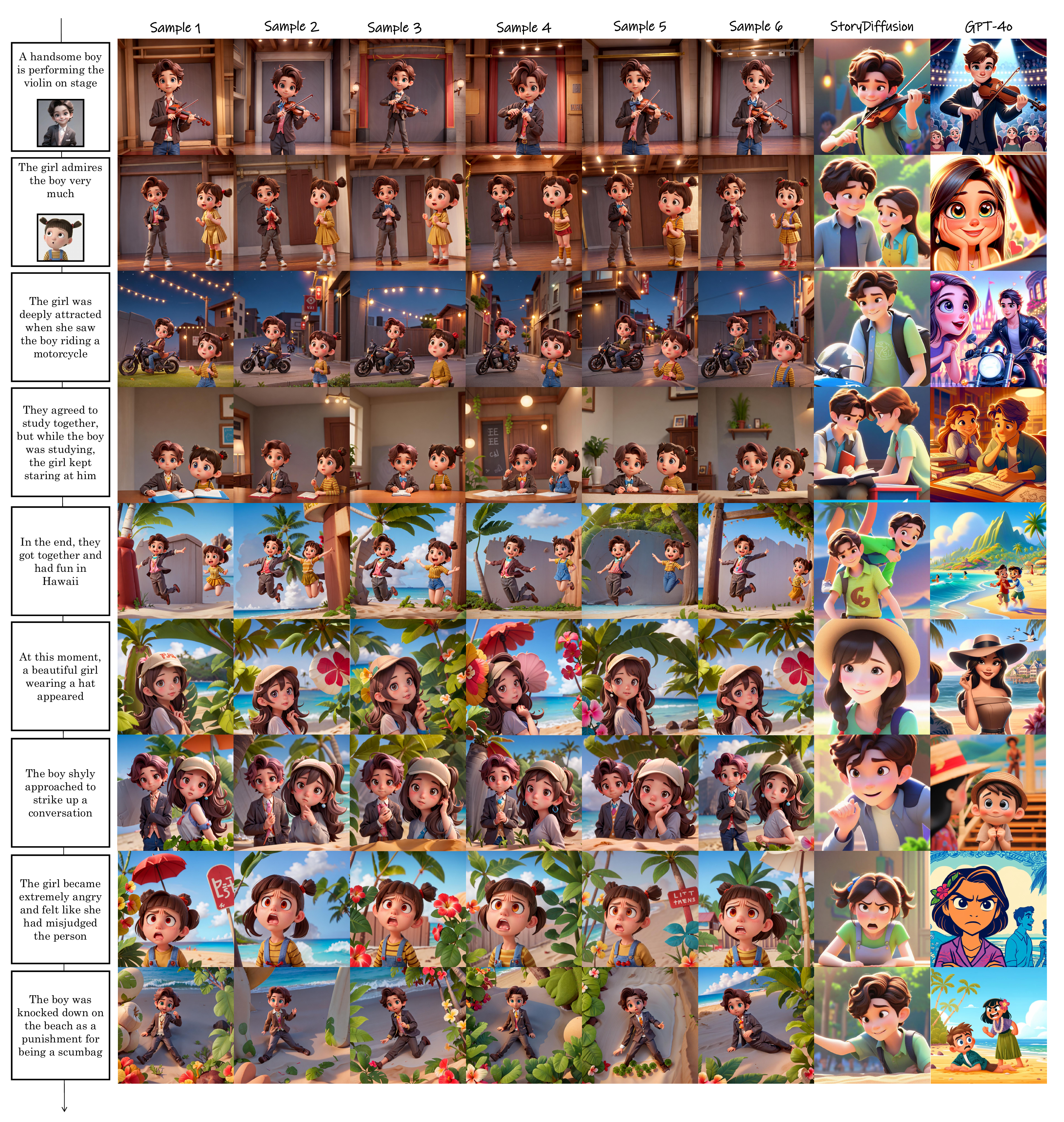}
  \caption{Multi-turn Interactive Image Generation results.}
  \label{fig: cjh2}
\end{figure*}

In Figure~\ref{fig: cjh1} and Figure~\ref{fig: cjh2}, we present additional comparison results between AutoStudio and existing methods in multi-turn interactive image generation. The results clearly demonstrate that AutoStudio outperforms other methods by better aligning with human needs. It enables the dynamic maintenance of consistent main characters while consistently producing high-quality images that meet user expectations.

\section{Limitations and Social Impacts}
\label{appn: Limitations and Social Impacts}

\paragraph{Limitations} Due to the inherent limitations of the T2I model (SD) itself, AutoStudio may exhibit abruptness when generating details especially in closely interactive scenarios between characters (such as hugging or lying in each other's arms). There is a possibility of abrupt effect leading to the generation of multiple hands or legs. Additionally, the involvement of multiple agents in the conversation might result in a slight increase in computational time and resource requirements. 

\paragraph{Broader Impacts} The generated content is influenced by the user's intentions, which we cannot control due to the interactive nature of AutoStudio. Harmful and explicit content such as pornography, violence, or graphic imagery can arise. However, this can be addressed by deploying a safety-finetuned version of SD, as AutoStudio is traning-free and flexible.
\newpage
\newpage
\newpage
\newpage
\newpage
\newpage
\newpage
\newpage
\newpage

\clearpage

\bibliographystyle{plain}
\bibliography{auto}

\begin{thebibliography}{10}

\bibitem{anonymous2024visdiahalbench}
Anonymous.
\newblock Visdiahalbench: A visual dialogue benchmark for diagnosing hallucination in large vision-language models.
\newblock In {\em The 62nd Annual Meeting of the Association for Computational Linguistics}, 2024.

\bibitem{avrahami2023chosen}
Omri Avrahami, Amir Hertz, Yael Vinker, Moab Arar, Shlomi Fruchter, Ohad Fried, Daniel Cohen-Or, and Dani Lischinski.
\newblock The chosen one: Consistent characters in text-to-image diffusion models.
\newblock {\em arXiv preprint arXiv:2311.10093}, 2023.

\bibitem{DALL·E3}
James Betker, Gabriel Goh, Li~Jing, Tim Brooks, Jianfeng Wang, Linjie Li, Long Ouyang, Juntang Zhuang, Joyce Lee, Yufei Guo, et~al.
\newblock Improving image generation with better captions.
\newblock {\em Computer Science. https://cdn. openai. com/papers/dall-e-3. pdf}, 2(3):8, 2023.

\bibitem{cheng2025animegamer}
Junhao Cheng, Yuying Ge, Yixiao Ge, Jing Liao, and Ying Shan.
\newblock Animegamer: Infinite anime life simulation with next game state prediction.
\newblock {\em arXiv preprint arXiv:2504.01014}, 2025.

\bibitem{cheng2024theatergen}
Junhao Cheng, Baiqiao Yin, Kaixin Cai, Minbin Huang, Hanhui Li, Yuxin He, Xi~Lu, Yue Li, Yifei Li, Yuhao Cheng, et~al.
\newblock Theatergen: Character management with llm for consistent multi-turn image generation.
\newblock {\em arXiv preprint arXiv:2404.18919}, 2024.

\bibitem{SEED}
Yuying Ge, Sijie Zhao, Ziyun Zeng, Yixiao Ge, Chen Li, Xintao Wang, and Ying Shan.
\newblock Making llama see and draw with seed tokenizer.
\newblock {\em arXiv preprint arXiv:2310.01218}, 2023.

\bibitem{TaleCrafter}
Yuan Gong, Youxin Pang, Xiaodong Cun, Menghan Xia, Haoxin Chen, Longyue Wang, Yong Zhang, Xintao Wang, Ying Shan, and Yujiu Yang.
\newblock Talecrafter: Interactive story visualization with multiple characters.
\newblock {\em arXiv preprint arXiv:2305.18247}, 2023.

\bibitem{GAN}
Ian Goodfellow, Jean Pouget-Abadie, Mehdi Mirza, Bing Xu, David Warde-Farley, Sherjil Ozair, Aaron Courville, and Yoshua Bengio.
\newblock Generative adversarial networks.
\newblock {\em Communications of the ACM}, pages 139--144, 2020.

\bibitem{gu2024mix}
Yuchao Gu, Xintao Wang, Jay~Zhangjie Wu, Yujun Shi, Yunpeng Chen, Zihan Fan, Wuyou Xiao, Rui Zhao, Shuning Chang, Weijia Wu, et~al.
\newblock Mix-of-show: Decentralized low-rank adaptation for multi-concept customization of diffusion models.
\newblock {\em Advances in Neural Information Processing Systems}, 36, 2024.

\bibitem{Guo_Chen_Wang}
Taicheng Guo, Xiuying Chen, Yaqi Wang, Ruidi Chang, Shichao Pei, NiteshV Chawla, Olaf Wiest, and Xiangliang Zhang.
\newblock Large language model based multi-agents: A survey of progress and challenges.

\bibitem{ddpm}
Jonathan Ho, Ajay Jain, and Pieter Abbeel.
\newblock Denoising diffusion probabilistic models.
\newblock {\em Advances in neural information processing systems}, 33:6840--6851, 2020.

\bibitem{Hong_Zheng_}
Sirui Hong, Xiawu Zheng, Jonathan Chen, Yuheng Cheng, Ceyao Zhang, Zili Wang, StevenKaShing Yau, Zijuan Lin, Liyang Zhou, Chenyu Ran, Lingfeng Xiao, and Chenglin Wu.
\newblock Metagpt: Meta programming for multi-agent collaborative framework.
\newblock Aug 2023.

\bibitem{hu2021lora}
Edward~J Hu, Yelong Shen, Phillip Wallis, Zeyuan Allen-Zhu, Yuanzhi Li, Shean Wang, Lu~Wang, and Weizhu Chen.
\newblock Lora: Low-rank adaptation of large language models.
\newblock {\em arXiv preprint arXiv:2106.09685}, 2021.

\bibitem{huang2024dialoggen}
Minbin Huang, Yanxin Long, Xinchi Deng, Ruihang Chu, Jiangfeng Xiong, Xiaodan Liang, Hong Cheng, Qinglin Lu, and Wei Liu.
\newblock Dialoggen: Multi-modal interactive dialogue system for multi-turn text-to-image generation.
\newblock {\em arXiv preprint arXiv:2403.08857}, 2024.

\bibitem{GANsu}
Minguk Kang, Jun-Yan Zhu, Richard Zhang, Jaesik Park, Eli Shechtman, Sylvain Paris, and Taesung Park.
\newblock Scaling up gans for text-to-image synthesis.
\newblock In {\em Proceedings of the IEEE/CVF Conference on Computer Vision and Pattern Recognition}, pages 10124--10134, 2023.

\bibitem{styleGAN}
Tero Karras, Samuli Laine, and Timo Aila.
\newblock A style-based generator architecture for generative adversarial networks.
\newblock In {\em Proceedings of the IEEE/CVF conference on computer vision and pattern recognition}, pages 4401--4410, 2019.

\bibitem{styleGAN2}
Tero Karras, Samuli Laine, Miika Aittala, Janne Hellsten, Jaakko Lehtinen, and Timo Aila.
\newblock Analyzing and improving the image quality of stylegan.
\newblock In {\em Proceedings of the IEEE/CVF conference on computer vision and pattern recognition}, pages 8110--8119, 2020.

\bibitem{SAM}
Alexander Kirillov, Eric Mintun, Nikhila Ravi, Hanzi Mao, Chloe Rolland, Laura Gustafson, Tete Xiao, Spencer Whitehead, Alexander~C Berg, Wan-Yen Lo, et~al.
\newblock Segment anything.
\newblock {\em arXiv preprint arXiv:2304.02643}, 2023.

\bibitem{li2023gligen}
Yuheng Li, Haotian Liu, Qingyang Wu, Fangzhou Mu, Jianwei Yang, Jianfeng Gao, Chunyuan Li, and Yong~Jae Lee.
\newblock Gligen: Open-set grounded text-to-image generation.
\newblock In {\em Proceedings of the IEEE/CVF Conference on Computer Vision and Pattern Recognition}, pages 22511--22521, 2023.

\bibitem{lian2023llm}
Long Lian, Boyi Li, Adam Yala, and Trevor Darrell.
\newblock Llm-grounded diffusion: Enhancing prompt understanding of text-to-image diffusion models with large language models.
\newblock {\em arXiv preprint arXiv:2305.13655}, 2023.

\bibitem{lin2024layoutprompter}
Jiawei Lin, Jiaqi Guo, Shizhao Sun, Zijiang Yang, Jian-Guang Lou, and Dongmei Zhang.
\newblock Layoutprompter: Awaken the design ability of large language models.
\newblock {\em Advances in Neural Information Processing Systems}, 36, 2024.

\bibitem{liu2023intelligent}
Chang Liu, Haoning Wu, Yujie Zhong, Xiaoyun Zhang, and Weidi Xie.
\newblock Intelligent grimm--open-ended visual storytelling via latent diffusion models.
\newblock {\em arXiv preprint arXiv:2306.00973}, 2023.

\bibitem{GroundingDino}
Shilong Liu, Zhaoyang Zeng, Tianhe Ren, Feng Li, Hao Zhang, Jie Yang, Chunyuan Li, Jianwei Yang, Hang Su, Jun Zhu, et~al.
\newblock Grounding dino: Marrying dino with grounded pre-training for open-set object detection.
\newblock {\em arXiv preprint arXiv:2303.05499}, 2023.

\bibitem{lu2020structured}
You Lu and Bert Huang.
\newblock Structured output learning with conditional generative flows.
\newblock In {\em Proceedings of the AAAI Conference on Artificial Intelligence}, volume~34, pages 5005--5012, 2020.

\bibitem{luo2025object}
Xiangyang Luo, Junhao Cheng, Yifan Xie, Xin Zhang, Tao Feng, Zhou Liu, Fei Ma, and Fei Yu.
\newblock Object isolated attention for consistent story visualization.
\newblock {\em arXiv preprint arXiv:2503.23353}, 2025.

\bibitem{DALLE3}
OpenAI.
\newblock Dall·e 3 system card.
\newblock 2023.

\bibitem{gpt4o}
OpenAI.
\newblock Gpt-4o.
\newblock 2024.

\bibitem{ostashev2024moa}
Daniil Ostashev, Yuwei Fang, Sergey Tulyakov, Kfir Aberman, et~al.
\newblock Moa: Mixture-of-attention for subject-context disentanglement in personalized image generation.
\newblock {\em arXiv preprint arXiv:2404.11565}, 2024.

\bibitem{Park_O}
JoonSung Park, JosephC. O’Brien, CarrieJ. Cai, MeredithRingel Morris, Percy Liang, and MichaelS. Bernstein.
\newblock Generative agents: Interactive simulacra of human behavior.
\newblock Apr 2023.

\bibitem{sdxl}
Dustin Podell, Zion English, Kyle Lacey, Andreas Blattmann, Tim Dockhorn, Jonas M{\"u}ller, Joe Penna, and Robin Rombach.
\newblock Sdxl: Improving latent diffusion models for high-resolution image synthesis.
\newblock {\em arXiv preprint arXiv:2307.01952}, 2023.

\bibitem{Qian_Cong_Yang_Chen_Su_Xu_Liu_Sun_2023}
Chen Qian, Xin Cong, Cheng Yang, Weize Chen, Yusheng Su, Juyuan Xu, Zhiyuan Liu, and Maosong Sun.
\newblock Communicative agents for software development.
\newblock Jul 2023.

\bibitem{Clip}
Alec Radford, Jong~Wook Kim, Chris Hallacy, Aditya Ramesh, Gabriel Goh, Sandhini Agarwal, Girish Sastry, Amanda Askell, Pamela Mishkin, Jack Clark, et~al.
\newblock Learning transferable visual models from natural language supervision.
\newblock In {\em International conference on machine learning}, pages 8748--8763, 2021.

\bibitem{DALL·E2}
Aditya Ramesh, Prafulla Dhariwal, Alex Nichol, Casey Chu, and Mark Chen.
\newblock Hierarchical text-conditional image generation with clip latents.
\newblock {\em arXiv preprint arXiv:2204.06125}, 2022.

\bibitem{sd}
Robin Rombach, Andreas Blattmann, Dominik Lorenz, Patrick Esser, and Bj{\"o}rn Ommer.
\newblock High-resolution image synthesis with latent diffusion models.
\newblock In {\em Proceedings of the IEEE/CVF conference on computer vision and pattern recognition}, pages 10684--10695, 2022.

\bibitem{sun2024spatial}
Wenqiang Sun, Teng Li, Zehong Lin, and Jun Zhang.
\newblock Spatial-aware latent initialization for controllable image generation.
\newblock {\em arXiv preprint arXiv:2401.16157}, 2024.

\bibitem{tang2024prioritizing}
Xiangru Tang, Qiao Jin, Kunlun Zhu, Tongxin Yuan, Yichi Zhang, Wangchunshu Zhou, Meng Qu, Yilun Zhao, Jian Tang, Zhuosheng Zhang, et~al.
\newblock Prioritizing safeguarding over autonomy: Risks of llm agents for science.
\newblock {\em arXiv preprint arXiv:2402.04247}, 2024.

\bibitem{team2023gemini}
Gemini Team, Rohan Anil, Sebastian Borgeaud, Yonghui Wu, Jean-Baptiste Alayrac, Jiahui Yu, Radu Soricut, Johan Schalkwyk, Andrew~M Dai, Anja Hauth, et~al.
\newblock Gemini: a family of highly capable multimodal models.
\newblock {\em arXiv preprint arXiv:2312.11805}, 2023.

\bibitem{tibebu2022text}
Haileleol Tibebu, Aadin Malik, and Varuna De~Silva.
\newblock Text to image synthesis using stacked conditional variational autoencoders and conditional generative adversarial networks.
\newblock In {\em Science and Information Conference}, pages 560--580, 2022.

\bibitem{AutoStory}
Wen Wang, Canyu Zhao, Hao Chen, Zhekai Chen, Kecheng Zheng, and Chunhua Shen.
\newblock Autostory: Generating diverse storytelling images with minimal human effort.
\newblock {\em arXiv preprint arXiv:2311.11243}, 2023.

\bibitem{wei2022chain}
Jason Wei, Xuezhi Wang, Dale Schuurmans, Maarten Bosma, Fei Xia, Ed~Chi, Quoc~V Le, Denny Zhou, et~al.
\newblock Chain-of-thought prompting elicits reasoning in large language models.
\newblock {\em Advances in neural information processing systems}, 35:24824--24837, 2022.

\bibitem{wu2023autogen}
Qingyun Wu, Gagan Bansal, Jieyu Zhang, Yiran Wu, Shaokun Zhang, Erkang Zhu, Beibin Li, Li~Jiang, Xiaoyun Zhang, and Chi Wang.
\newblock Autogen: Enabling next-gen llm applications via multi-agent conversation framework.
\newblock {\em arXiv preprint arXiv:2308.08155}, 2023.

\bibitem{yang2024mastering}
Ling Yang, Zhaochen Yu, Chenlin Meng, Minkai Xu, Stefano Ermon, and Bin Cui.
\newblock Mastering text-to-image diffusion: Recaptioning, planning, and generating with multimodal llms.
\newblock {\em arXiv preprint arXiv:2401.11708}, 2024.

\bibitem{ipadapter}
Hu~Ye, Jun Zhang, Sibo Liu, Xiao Han, and Wei Yang.
\newblock Ip-adapter: Text compatible image prompt adapter for text-to-image diffusion models.
\newblock {\em arXiv preprint arXiv:2308.06721}, 2023.

\bibitem{Mora}
Zhengqing Yuan, Ruoxi Chen, Zhaoxu Li, Haolong Jia, Lifang He, Chi Wang, Lichao Sun, and /~Mora.
\newblock Mora: Enabling generalist video generation via a multi-agent framework.

\bibitem{Mini-DALLE3}
Lai Zeqiang, Zhu Xizhou, Dai Jifeng, Qiao Yu, and Wang Wenhai.
\newblock Mini-dalle3: Interactive text to image by prompting large language models.
\newblock {\em arXiv preprint arXiv:2310.07653}, 2023.

\bibitem{controlnet}
Lvmin Zhang, Anyi Rao, and Maneesh Agrawala.
\newblock Adding conditional control to text-to-image diffusion models.
\newblock In {\em Proceedings of the IEEE/CVF International Conference on Computer Vision}, pages 3836--3847, 2023.

\bibitem{zheng2023layoutdiffusion}
Guangcong Zheng, Xianpan Zhou, Xuewei Li, Zhongang Qi, Ying Shan, and Xi~Li.
\newblock Layoutdiffusion: Controllable diffusion model for layout-to-image generation.
\newblock In {\em Proceedings of the IEEE/CVF Conference on Computer Vision and Pattern Recognition}, pages 22490--22499, 2023.

\bibitem{MiniGPT5}
Kaizhi Zheng, Xuehai He, and Xin~Eric Wang.
\newblock Minigpt-5: Interleaved vision-and-language generation via generative vokens.
\newblock {\em arXiv preprint arXiv:2310.02239}, 2023.

\bibitem{zhou2024storydiffusion}
Yupeng Zhou, Daquan Zhou, Ming-Ming Cheng, Jiashi Feng, and Qibin Hou.
\newblock Storydiffusion: Consistent self-attention for long-range image and video generation.
\newblock {\em arXiv preprint arXiv:2405.01434}, 2024.

\end{thebibliography}

\end{document}